\PassOptionsToPackage{dvipsnames}{xcolor}

\documentclass{article} %
\usepackage{iclr2023_conference,times}
\iclrfinalcopy

\usepackage{amsmath,amsfonts,bm}

\def\eqref#1{equation~\ref{#1}}

\def\1{\bm{1}}

\DeclareMathAlphabet{\mathsfit}{\encodingdefault}{\sfdefault}{m}{sl}
\SetMathAlphabet{\mathsfit}{bold}{\encodingdefault}{\sfdefault}{bx}{n}

\newcommand{\D}{\mathcal{D}}

\usepackage{hyperref}
\usepackage{url}
\usepackage{framed} 
\usepackage{multirow}
\usepackage{booktabs}
\usepackage{xspace}
\usepackage{caption}
\usepackage{subcaption}
\usepackage{graphicx}
\newcommand{\algname}{RT-1\xspace}
\newcommand{\algfullname}{Robotics Transformer 1\xspace}

\graphicspath{{figures/}}
\usepackage{array}
\usepackage{enumitem}

\usepackage{floatrow}
\newfloatcommand{capbtabbox}{table}[][\FBwidth]

\newcolumntype{P}[1]{>{\centering\arraybackslash}p{#1}}

\title{RT-1: Robotics Transformer \\for Real-World Control at Scale}

\begin{document}

\maketitle

\vspace{-2cm}
{
\begin{flushleft}
\footnote{Authors listed in alphabetical order. Contributions in Appendix~\ref{sec:app_contributions}.\newline Corresponding emails: \texttt{\{keerthanapg,kanishkarao,karolhausman\}@google.com}.}
\bf\small\mbox{Anthony Brohan$^*$}, \mbox{Noah Brown$^*$},  \mbox{Justice Carbajal$^*$}, \mbox{Yevgen Chebotar$^*$}, \mbox{Joseph Dabis$^*$}, \mbox{Chelsea Finn$^*$}, \mbox{Keerthana Gopalakrishnan$^*$}, \mbox{Karol Hausman$^*$}, \mbox{Alex Herzog$^\dag$}, \mbox{Jasmine Hsu$^*$}, \mbox{Julian Ibarz$^*$},
\mbox{Brian Ichter$^*$}, \mbox{Alex Irpan$^*$}, \mbox{Tomas Jackson$^*$}, 
\mbox{Sally Jesmonth$^*$}, \mbox{Nikhil J Joshi$^*$}, \mbox{Ryan Julian$^*$}, \mbox{Dmitry Kalashnikov$^*$}, \mbox{Yuheng Kuang$^*$}, \mbox{Isabel Leal$^*$}, \mbox{Kuang-Huei Lee$^\ddag$},
\mbox{Sergey Levine$^*$}, \mbox{Yao Lu$^*$}, \mbox{Utsav Malla$^*$}, \mbox{Deeksha Manjunath$^*$}, \mbox{Igor Mordatch$^\ddag$},  \mbox{Ofir Nachum$^\ddag$}, \mbox{Carolina Parada$^*$}, \mbox{Jodilyn Peralta$^*$}, \mbox{Emily Perez$^*$}, \mbox{Karl Pertsch$^*$}, \mbox{Jornell Quiambao$^*$},
\mbox{Kanishka Rao$^*$}, \mbox{Michael Ryoo$^*$}, \mbox{Grecia Salazar$^*$}, \mbox{Pannag Sanketi$^*$}, \mbox{Kevin Sayed$^*$}, \mbox{Jaspiar Singh$^*$}, \mbox{Sumedh Sontakke$^\ddag$}, \mbox{Austin Stone$^*$}, \mbox{Clayton Tan$^*$},
\mbox{Huong Tran$^*$}, \mbox{Vincent Vanhoucke$^*$}, \mbox{Steve Vega$^*$}, \mbox{Quan Vuong$^*$}, \mbox{Fei Xia$^*$}, \mbox{Ted Xiao$^*$}, \mbox{Peng Xu$^*$}, \mbox{Sichun Xu$^*$}, \mbox{Tianhe Yu$^*$}, \mbox{Brianna Zitkovich$^*$} 
\end{flushleft}
\vspace{-0.25cm}
$^*$\href{http://g.co/robotics}{Robotics at Google}, $^\dag$\href{https://everydayrobots.com/}{Everyday Robots},
$^\ddag$\href{https://research.google/teams/brain/}{Google Research, Brain Team}
}

\begin{abstract}
By transferring knowledge from large, diverse, task-agnostic datasets, modern machine learning models can solve specific downstream tasks either zero-shot or with small task-specific datasets to a high level of performance. 
While this capability has been demonstrated in other fields such as computer vision, natural language processing or speech recognition, it remains to be shown in robotics, where the generalization capabilities of the models are particularly critical due to the difficulty of collecting real-world robotic data.
We argue that one of the keys to the success of such general robotic models lies with open-ended task-agnostic training, combined with high-capacity architectures that can 
absorb all of the diverse, robotic data.
In this paper, we present a model class, dubbed Robotics Transformer, that exhibits promising scalable model properties.
We verify our conclusions in a study of different model classes and their ability to generalize as a function of the data size, model size, and data diversity based on a large-scale data collection on real robots performing real-world tasks.
The project’s website and videos can
be found at \url{robotics-transformer1.github.io}
\end{abstract}

\section{Introduction}\label{sec:intro}
\vspace{-0.2cm}
End-to-end robotic learning, with either imitation or reinforcement, typically involves collecting task-specific data in either single-task~\citep{kalashnikov2018qtopt, zhang2018deep} or multi-task~\citep{mtopt2021arxiv, jang2022bc} settings that are narrowly tailored to the tasks that the robot should perform. This workflow mirrors the classic approach to supervised learning in other domains, such as computer vision and NLP, where task-specific datasets would be collected, labeled, and deployed to solve individual tasks, with little interplay between the tasks themselves.
Recent years have seen a transformation in vision, NLP, and other domains, away from siloed, small-scale datasets and models and towards large, general models pre-trained on broad, large datasets. The keys to the success of such models lie with open-ended task-agnostic training, combined with high-capacity architectures that can absorb all of the knowledge present in large-scale datasets.
If a model can ``sponge up'' experience to learn general patterns in language or perception, then it can bring them to bear on individual tasks more efficiently.
While removing the need for large task-specific datasets is appealing generally in supervised learning, it is even more critical in robotics, where datasets might require engineering-heavy autonomous operation or expensive human demonstrations. We therefore ask: can we train a single, capable, large multi-task backbone model on data consisting of a wide variety of robotic tasks? And does such a model enjoy the benefits observed in other domains, exhibiting zero-shot generalization to new tasks, environments, and objects?

Building such models in robotics is not easy. Although recent years have seen several large multi-task robot policies proposed in the literature~\citep{reed2022generalist,jang2022bc}, such models
often have limited breadth of real-world tasks, as with Gato~\citep{reed2022generalist}, or focus on training tasks rather than generalization to new tasks, as with recent instruction following methods~\citep{shridhar2021cliport,shridhar2022perceiver}, or attain comparatively lower performance on new tasks~\citep{jang2022bc}.

\begin{figure}
     \centering
     \begin{subfigure}[b]{\textwidth}
         \centering
         \includegraphics[width=\textwidth]{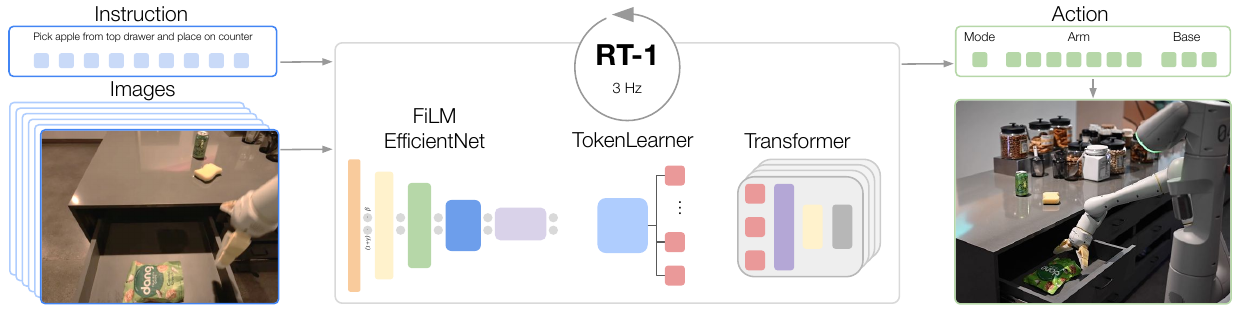}
         \caption{\algname takes images and natural language instructions and outputs discretized base and arm actions. Despite its size (35M parameters), it does this at 3~Hz, due to its efficient yet high-capacity architecture: a FiLM~\citep{Perez_Strub} conditioned EfficientNet~\citep{pmlr-v97-tan19a}, a TokenLearner~\citep{ryoo2021tokenlearner}, and a Transformer~\citep{vaswani2017attention}.
         }
\label{fig:intro_model}
     \end{subfigure}
     \hfill
     \begin{subfigure}[b]{\textwidth}
         \centering
         \includegraphics[width=0.99\textwidth]{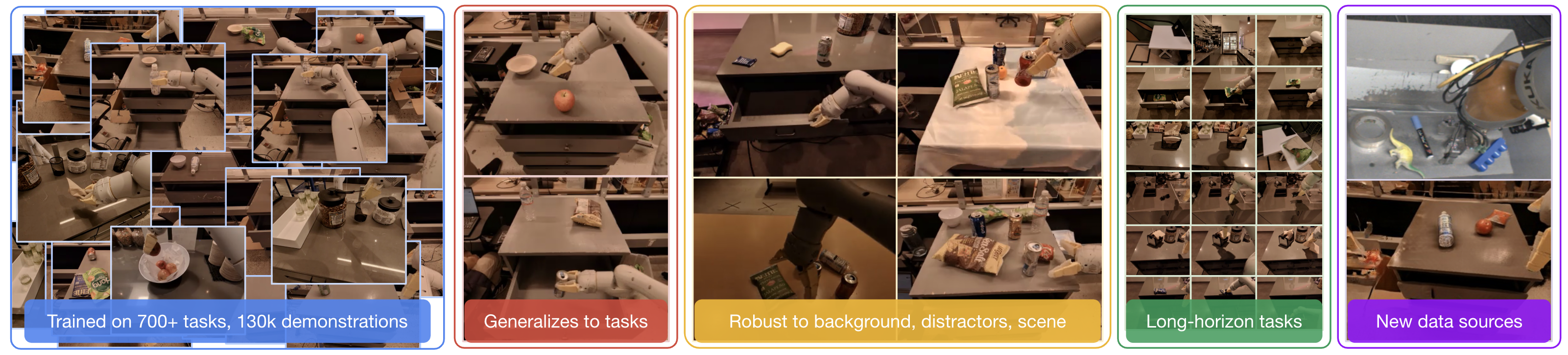}
         \caption{\algname's large-scale, real-world training (130k demonstrations) and evaluation (3000 real-world trials) show impressive generalization, robustness, and ability to learn from diverse data.}
         \label{fig:intro_tasks}
     \end{subfigure}
        \caption{A high-level overview of \algname's architecture, dataset, and evaluation.}
        \label{fig:intro}
\vspace{-0.2cm}
\end{figure}

The two main challenges lie in assembling the right dataset and designing the right model. While data collection and curation is often the ``unsung hero'' of many large-scale machine learning projects~\citep{radford2021learning,ramesh2021zero}, this is especially true in robotics, where datasets are often robot-specific and gathered manually~\citep{dasari2019robonet, ebert2021bridge}. As we will show in our evaluations, good generalization requires datasets that combine both scale and breadth, covering a variety of tasks and settings. At the same time, the tasks in the dataset should be sufficiently well-connected to enable generalization, such that the model can discover the patterns between structural similar tasks and perform new tasks that combine those patterns in novel ways.
We utilize a dataset that we gathered over the course of 17 months with a fleet of 13 robots, containing $\sim$130k episodes and over 700 tasks, and we ablate various aspects of this dataset in our evaluation.

The second challenge lies in the design of the model itself. Effective robotic multi-task learning requires a high capacity model, and Transformer~\citep{vaswani2017attention} models excel in this regard, particularly when it is necessary to learn many tasks conditioned, as in our case, on language instructions. However, robotic controllers must also be efficient enough to run in real time, which presents a major challenge for Transformers in particular. We propose a novel architecture that we call \algname (\algfullname), which 
by encoding high-dimensional inputs and outputs, including camera images, instructions and motor commands into compact token representations to be used by the Transformer,
allows for efficient inference at runtime to make real-time control feasible.

Our contribution is the RT-1 model and experiments with this model on a large and broad dataset of real-world robotic tasks. Our experiments not only demonstrate that RT-1 can exhibit significantly improved generalization and robustness compared to prior techniques, but also evaluate and ablate many design choices in both the model and in the composition of the training set.
Our results show that \algname can perform over 700 training instructions at 97\% success rate, and can generalize to new tasks, distractors, and backgrounds 25\%, 36\% and 18\% better than the next best baseline, respectively. This level of performance allows us to execute very long-horizon tasks in the SayCan~\citep{ahn2022can} framework, with as many as 50 stages. We further show that \algname can incorporate data from simulation or even other robot types, retaining performance on the original tasks and improving generalization to new scenarios. A short overview of \algname capabilities is presented in Fig.~\ref{fig:intro_tasks}\footnote{Helper robots shown in Fig. 1-5 are from \href{http://www.everydayrobots.com}{Everyday Robots}}.

\section{Related Work}\label{sec:related}

A number of recent works have proposed Transformer-based policies for robotic control. As in RT-1, several works use language commands processed with Transformers as a robust framework for specifying and generalizing to new tasks~\citep{zhang2021hierarchical,pashevich2021episodic,silva2021lancon,jang2022bc,ahn2022can,nair2022learning}.
Our work takes the application of Transformers a step further and treats the mapping of language and vision observations to robot actions as a sequence modelling problem, using a Transformer to learn this mapping. This idea is directly inspired by successes in game-playing~\citep{chen2021decision,lee2022multi} as well as simulated robot navigation~\citep{fang2019scene}, locomotion~\citep{janner2021reinforcement,gupta2022metamorph}, and manipulation~\citep{jiang2022vima} environments.  We note that several of these works go beyond only text conditioning and use Transformers to also generalize across robot morphologies (e.g.,~\citet{gupta2022metamorph}) and other modalities for task specifications (e.g.,~\citet{jang2022bc,jiang2022vima}). These extensions are promising future directions for RT-1.

Beyond Transformer-based policies, the focus of our work is on generalizable and robust real-world robotic manipulation at scale.
Existing works on real-world Transformer-based robotic manipulation focus on efficiently learning tasks from a set of demonstrations per task~\citep{shridhar2022perceiver}. Behavior Transformer~\citep{shafiullah2022behavior} and Gato~\citep{reed2022generalist} advocate for training a single model on large-scale robotic and non-robotic datasets. However, these works are limited in their real-world robotic tasks; e.g., Gato learns effectively a single task (colored block stacking) without evaluating generalization to new tasks or a variety of real-world settings. On the technical side, our work examines how Transformer-based policies can be built so as to combine high capacity and generalization with the computational efficiency necessary for real-time control.

While the use of high-capacity Transformer models to learn robotic control policies is a fairly recent innovation, robotics has a long history of multi-task and language-conditioned learning, and RT-1 builds on these foundations. 
A significant body of work deals with learning policies and predictive models for robotic grasping~\citep{saxena2006robotic,lenz2015deep,pinto2016supersizing,gupta2018robot,viereck2017learning}, with the aim of generalizing to new objects. 
Prior works have sought to address robotic language understanding through pipelined approaches that combine language parsing, vision, and robotic control~\citep{macmahon2006walk,kollar2010toward,tellex2011understanding} and with end-to-end approaches~\citep{mei2016listen,stepputtis2020language,lynch2020language,ahn2022can}. 
Multi-task robotic learning has also been approached from the perspective of learning to reach goals~\citep{chung2015bayesian,raffin2019decoupling,jurgenson2020sub,huang2020motion}, as well as learning policies that can perform tasks in a discrete set or some other parameterized form~\citep{deisenroth2014multi,devin2017learning,fox2019multi,kalashnikov2021mt}. 
A number of prior works in robotics have also focused on collecting datasets containing demonstrations or trials that illustrate a variety of different tasks~\citep{sharma2018multiple,dasari2019robonet,yu2020meta,singh2020scalable,james2020rlbench}.
Our work adds further evidence in support of the power of multi-task, language-conditioned robotic learning, presenting experimental results at a larger scale and with a greater variety of behaviors, objects, and scenes and proposing new architectures and design choices that enable robotic learning at a significantly larger scale.

\section{Preliminaries}\label{sec:prelims}

\paragraph{Robot learning.} We aim to learn robot policies to solve language-conditioned tasks from vision. Formally, we consider a sequential decision-making environment. At timestep $t=0$, the policy $\pi$ is presented with a language instruction $i$ and an initial image observation $x_0$. The policy produces an action distribution $\pi(\cdot~|~i,x_0)$ from which an action $a_0$ is sampled and applied to the robot. This process continues, with the policy iteratively producing actions $a_t$ by sampling from a learned distribution $\pi(\cdot~|~i,\{x_j\}_{j=0}^t)$ and applying those actions to the robot. The interaction ends when a termination condition is achieved. %
The full interaction $i,\{(x_j,a_j)\}_{j=0}^T$ from from the starting step $t=0$ to terminating step $T$ is referred to as an \emph{episode}.
At the end of an episode, the agent will be given a binary reward $r\in\{0,1\}$ indicating whether the robot performed the instruction $i$. The goal is to learn a policy $\pi$ that maximizes the average reward, in expectation over a distribution of instructions, starting states $x_0$, and transition dynamics.

\noindent\textbf{Transformers.}\label{sec:prelims_transformers}
RT-1 uses a Transformer~\citep{vaswani2017attention} to parameterize the policy $\pi$. Generally speaking, a Transformer is a sequence model mapping an input sequence $\{\xi_h\}_{h=0}^H$ to an output sequence $\{y_k\}_{k=0}^K$ using combinations of self-attention layers and fully-connected neural networks. While Transformers were originally designed for text sequences, where each input $\xi_j$ and output $y_k$ represents a text token, they have been extended to images~\citep{parmar2018image} as well as other modalities~\citep{lee2022multi,reed2022generalist}.
As detailed in the next section, we parameterize $\pi$ by first mapping inputs $i,\{x_j\}_{j=0}^t$ to a sequence $\{\xi_h\}_{h=0}^H$ and action outputs $a_t$ to a sequence $\{y_k\}_{k=0}^K$ before using a Transformer to learn the mapping $\{\xi_h\}_{h=0}^H\to\{y_k\}_{k=0}^K$.

\noindent\textbf{Imitation learning.}\label{sec:prelims_il}
Imitation learning methods train the policy $\pi$ on a dataset $\D$ of demonstrations~\citep{pomerleau1988alvinn,zhang2018deep,jang2022bc}. Specifically, we assume access to a dataset $\D=\{(i^{(n)}, \{(x_t^{(n)},a_t^{(n)})\}_{t=0}^{T^{(n)}})\}_{n=0}^N$ of episodes, all of which are successful (i.e., have a final reward of $1$). 
We learn $\pi$ using \emph{behavioral cloning}~\citep{pomerleau1988alvinn}, which optimizes $\pi$ by minimizing the negative log-likelihood of actions $a_t$ given the images and language instructions.

\section{System Overview}
\label{sec:system}
\begin{figure}[h]
     \centering
     \includegraphics[width=0.8\textwidth]{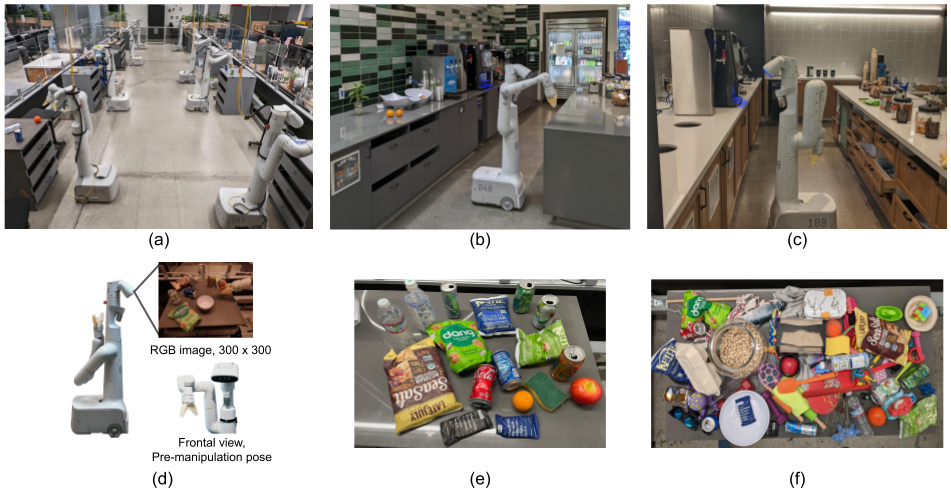}
     \caption{
{(a) Robot classroom where we collect data at scale; (b) a real office kitchen, one of the two realistic environments used for evaluation (named Kitchen1 in the rest of the paper); (c) a different office kitchen used for evaluation (named Kitchen2 in the rest of the paper); (d) mobile manipulator used throughout the paper; (e) a set of objects used for most of the skills to expand skill diversity; (f) a more diverse set of objects used mostly to expand object diversity of the picking skill.}
     }\label{fig:robot_setup}
\end{figure}

The goal of this work is to build and demonstrate a general robot learning system that can absorb large amounts of data and generalize effectively.
We use mobile manipulators from Everyday Robots\footnote{\url{everydayrobots.com}}, which have a 7 degree-of-freedom arm, a two-fingered gripper, and a mobile base (see Fig.~\ref{fig:robot_setup} (d)). To collect data and evaluate our method, we use three kitchen-based environments: two real office kitchens and a training environment modelled off these real kitchens. 
The training environment, shown in Fig.~\ref{fig:robot_setup} (a), consists of partial counters and is constructed for large scale data collection.
The two real environments, shown in Fig.~\ref{fig:robot_setup} (b, c), have similar counter tops to the training environment, but vary in lighting, background, and full kitchen geometry (e.g., there may be a cabinet instead of a drawer or a sink may be visible).
We evaluate the performance of our policies across these different environments, measuring the policy's performance and ability to generalize.

Our training data consists of human-provided demonstrations, and we annotate each episode with a textual description of the instruction that the robot just performed.  
The instructions usually contain a verb and one or more nouns describing the target objects. 
To group these instructions together, we split them into a number of skills (e.g., verbs such as ``pick'', ``open'' or ``place upright'') and objects (e.g., nouns such as ``coke can'', ``apple'', or ``drawer''). We describe the details of our data collection strategy at scale in Sec.~\ref{sec:data}.
Our largest dataset contains over 130k individual demonstrations constituting over 700 distinct task instructions using a large variety of objects (see Fig.~\ref{fig:robot_setup} (f)). We describe the details of the data collected in Sec.~\ref{sec:data}.

One of the main contributions of our system is the network architecture, \algfullname (\algname), an efficient model that can absorb large amounts of data,  effectively generalize, and output actions at real-time rates for practical robotic control. \algname takes a short sequence of images and a natural language instruction as input and outputs an action for the robot at each time step.
To this end, the architecture (shown in Figure~\ref{fig:intro_model}) leverages several elements: first the images and text are processed via an ImageNet pretrained convolutional network~\citep{pmlr-v97-tan19a} conditioned on a pretrained embedding of the instruction via FiLM~\citep{Perez_Strub}, followed by a Token Learner~\citep{ryoo2021tokenlearner} to compute a compact set of tokens, and finally a Transformer~\citep{vaswani2017attention} to attend over these tokens and produce discretized action tokens.
The actions consist of seven dimensions for the arm movement (x, y, z, roll, pitch, yaw, opening of the gripper), three dimensions for base movement (x, y, yaw) and a discrete dimension to switch between three modes: controlling the arm, the base, or terminating the episode.
\algname performs closed-loop control and commands actions at $3$~Hz until it either yields a ``terminate'' action or hits a pre-set time step limit.

\section{\algname: Robotics Transformer}\label{sec:robotics_transformer}

In this section, we describe how we tokenize the images, text, and actions, and then discuss the \algname model architecture. 
We then describe how we attain the runtime speed required for real-time control.
Lastly, we describe the data collection procedure and the skills and instructions in our dataset.

\begin{figure}[h]
     \centering
     \includegraphics[width=0.6\textwidth]{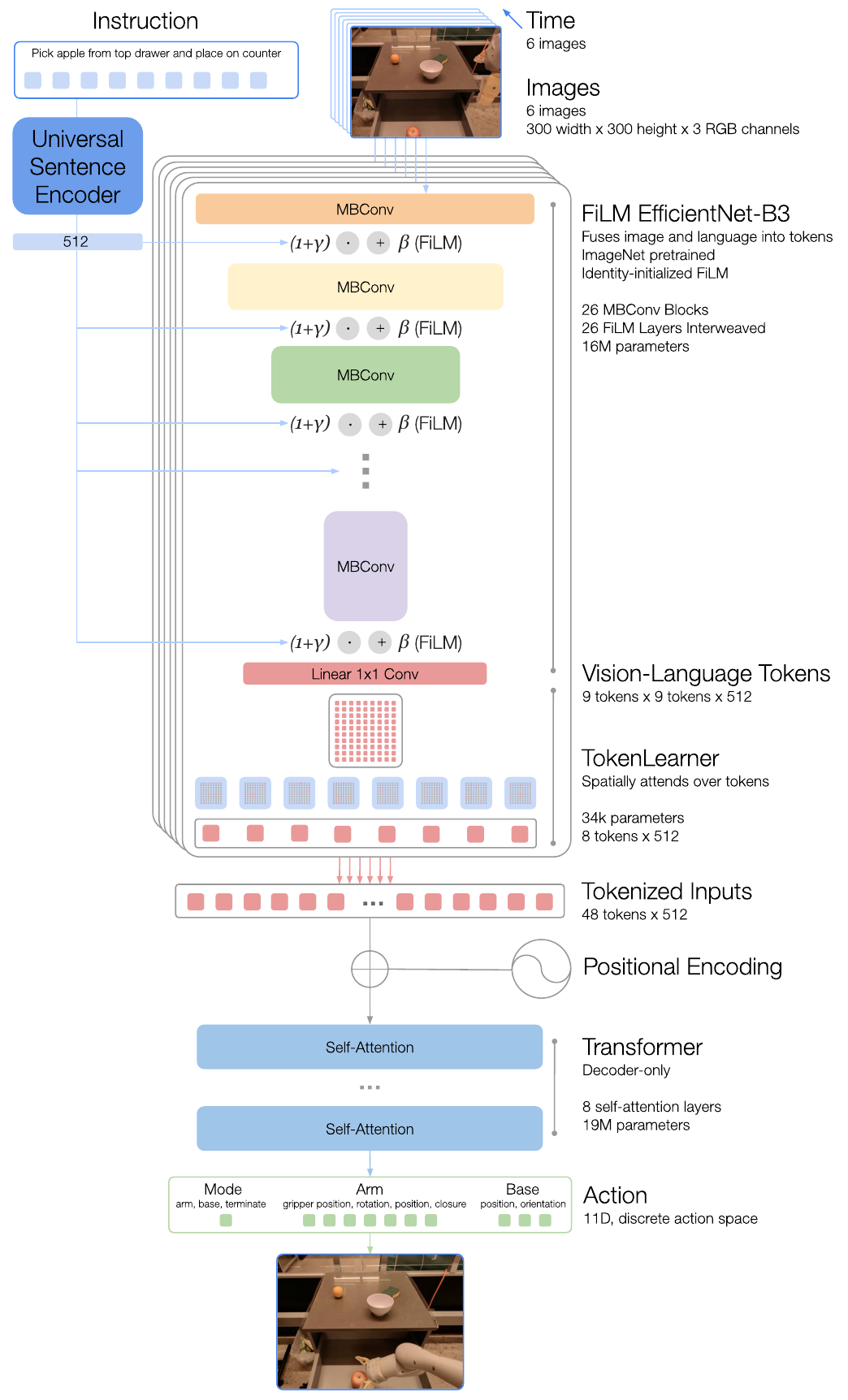}
     \caption{
{The architecture diagram of \algname. The instruction is transformed into a USE embedding and used to condition a pre-trained EfficientNet via FiLM layers. The resulting vision-language tokens are reduced by the TokenLearner and fed into a decoder-only Transformer, which outputs tokenized actions.}
     }\label{fig:model_full}
\end{figure}

\subsection{Model}\label{sec:model}
Our model is built on a Transformer architecture~\citep{vaswani2017attention} and takes a history of images and task description as input and directly outputs tokenized actions, as shown in Fig.~\ref{fig:intro_model} and in detail in Fig.~\ref{fig:model_full}.
In the following we describe the components of the model, following the top-to-bottom order in Fig.~\ref{fig:model_full}.
More detail on model selection at scale are provided in Appendix~\ref{sec:app_model_selection}.

\textbf{Instruction and image tokenization.} The \algname architecture relies on a data-efficient and compact tokenization of images \textit{and} language instruction. 
\algname tokenizes a history of 6 images by passing images through an ImageNet pretrained EfficientNet-B3 \citep{pmlr-v97-tan19a} model, which takes 6 images of resolution $300 \times 300$ as input and outputs a spatial feature map of shape $9 \times 9 \times 512$ from the final convolutional layer. 
Unlike~\citet{reed2022generalist}, we do not patchify the images into visual tokens prior to feeding them to our Transformer backbone. 
We instead flatten the output feature map from the EfficientNet into $81$ visual tokens which are passed on to the later layers of the network.

To include the language instruction, we condition the image tokenizer on the natural language instruction in the form of a pretrained language embedding, allowing extraction of task-relevant image features early on and improving performance of \algname. 
The instruction is first embedded via Universal Sentence Encoder~\citep{cer2018universal}. 
This embedding is then used as input to identity-initialized FiLM layers \citep{Perez_Strub} added to the pretrained EfficientNet to condition the image encoder. Normally, inserting a FiLM layer into the interior of a pretrained network would disrupt the intermediate activations and negate the benefit of using pretrained weights. To overcome this, we initialize the weights of the dense layers ($f_{c}$ and $h_{C}$) which produce the FiLM affine transformation to zero, allowing the FiLM layer to initially act as an identity and preserve the function of the pretrained weights. We find that identity-initialized FiLM also produces better results when training with an EfficientNet initialized from scratch, without ImageNet pretraining, but it does not surpass the initialization described above. The architecture of the image tokenizer is presented in Fig.~\ref{fig:model_full}.

\algname's image and instruction tokenization via FiLM EfficientNet-B3 is a total of 16M parameters, with 26 layers of MBConv blocks and FiLM layers, which output 81 vision-language tokens.

\textbf{TokenLearner.}
To further compress the number of tokens that \algname needs to attend over and thus speed up inference, \algname uses TokenLearner~\citep{ryoo2021tokenlearner}.
TokenLearner is an element-wise attention module that learns to map a large number of tokens into a much smaller number of tokens. This allows us to soft-select image tokens based on their information, passing only the important token combinations to the subsequent Transformer layers.
The inclusion of TokenLearner subsamples the $81$ visual tokens that come out of the pre-trained FiLM-EfficientNet layers to just $8$ final tokens that are then passed on to our Transformer layers.

\textbf{Transformer.}
These 8 tokens per-image are then concatenated with the other images in the history, forming 48 total tokens (with added position encoding) to be fed into the Transformer backbone of \algname.
The Transformer is a decoder-only sequence model with 8 self-attention layers and 19M total parameters that outputs action tokens.

\textbf{Action tokenization.}
To tokenize actions, each action dimension in \algname is discretized into $256$~bins. As mentioned previously, the action dimensions we consider include seven variables for the arm movement ($x$, $y$, $z$, roll, pitch, yaw, opening of the gripper), three variables for base movement ($x$, $y$, yaw) and a discrete variable to switch between three modes: controlling arm, base or terminating the episode.
For each variable, we map the target to one of the 256 bins, where the bins are uniformly distributed within the bounds of each variable.

\textbf{Loss.}
We use a standard categorical cross-entropy entropy objective and causal masking that was utilized in prior Transformer-based controllers~\citep{reed2022generalist, lee2022multi}.

\textbf{Inference speed.}
In contrast to many applications of large models, such as natural language or image generation, one of the unique requirements for a model that needs to run on real robots in real time is fast and consistent inference speed. 
Given the human speeds of executing the instructions considered in this work (which we measured to be in the $2-4$ secs range), we want the model to be not significantly slower than that. Based on our experiments this requirement corresponds to at least $3$Hz control frequency and the resulting inference time budget for the model, given other latencies in the system, to be less than $100$ms.

This requirement limits the size of the model that we can use. We further explore the impact of model size on inference speed in the experiments.
We employ two techniques to speed up inference: (i) reduce the number of tokens generated by a pre-trained EfficientNet model by using TokenLearner~\citep{ryoo2021tokenlearner}, (ii) compute these tokens only once and reuse them for the following windows that overlap for the future inferences. Both of these allow us to speed up the model inference by $2.4$ and $1.7$ times, respectively. 
Additional details on model inference are in Appendix~\ref{sec:app_inference}.

\subsection{Data}\label{sec:data}

\begin{table}[h!]
\begin{center}
\setlength\tabcolsep{2.0pt}
\scriptsize
\begin{tabular}{p{3.5cm} p{0.75cm} p{4.5cm} p{4cm}}
\toprule
Skill & Count & Description & Example Instruction\\
\midrule

Pick \texttt{Object} & 130 & Lift the object off the surface & pick iced tea can \\

Move \texttt{Object} Near \texttt{Object} & 337 & Move the first object near the second & move pepsi can near rxbar blueberry\\

Place \texttt{Object} Upright & 8 & Place an elongated object upright & place water bottle upright\\

Knock \texttt{Object} Over & 8 & Knock an elongated object over & knock redbull can over\\

Open \texttt{Drawer} & 3 & Open any of the cabinet drawers & open the top drawer \\

Close \texttt{Drawer}  & 3 & Close any of the cabinet drawers & close the middle drawer \\

Place \texttt{Object} into \texttt{Receptacle} & 84 & Place an object into a receptacle & place brown chip bag into white bowl\\

Pick \texttt{Object} from \texttt{Receptacle} and Place on the Counter & 162 & Pick an object up from a location and then place it on the counter  & pick green jalapeno chip bag from paper bowl and place on counter \\
\midrule
Section \ref{sec:exp_limit} and \ref{sec:exp_long} tasks & 9 & Skills trained for realistic, long instructions & open the large glass jar of pistachios \\
& & & pull napkin out of dispenser \\
& & & grab scooper \\

\midrule
Total & 744 \\
\bottomrule
\end{tabular}
\caption{The list of skills collected for \algname together with their descriptions and example instructions.}
\label{table:all_skills}
\end{center}
\end{table}

Our goal is to build a system that exhibits high performance, generalization to new tasks, and robustness to distractors and backgrounds. We therefore aim to collect a large, diverse dataset of robot trajectories that includes multiple tasks, objects and environments. 
Our primary dataset consists of $\sim$130k robot demonstrations, collected with a fleet of 13 robots over the course of 17 months.
We conducted this large-scale data collection in a series of office kitchen segments, which we refer to as \textit{robot classrooms}, shown in Fig.~\ref{fig:robot_setup}. More details on data collection are in Appendix~\ref{sec:app_collection}.

\textbf{Skills and instructions.} While the definition of a task remains inconsistent in the literature, in this work we count the number of language instructions that the system can perform, where an instruction corresponds to a verb surrounded by one or multiple nouns, such as ``\textit{place} water bottle \textit{upright}'', ``\textit{move} the coke can \textit{to} the green chip bag'' or ``\textit{open} the drawer''. 
\algname is able to perform  over 700 language instructions in multiple realistic office kitchen environments that we evaluate and describe in detail in the experiments. 
In order to group the evaluations and draw conclusions on the performance of the system, we group the instructions by the verbs used in them, which we refer to as \textit{skills}. 
A more detailed list of instructions is shown in Table~\ref{table:all_skills}, with examples and the number of instructions per skill. 

The current set of skills includes picking, placing, opening and closing drawers, getting items in and out drawers, placing elongated items up-right, knocking them over, pulling napkins and opening jars. 
The skills were chosen to demonstrate multiple behaviors with many objects (seen in Fig.~\ref{fig:robot_setup}(e)) to test aspects of \algname such as generalization to new instructions and ability to perform many tasks. 
We then greatly expanded the object diversity for the ``pick'' skill to make sure that the skills generalize to varied objects (see the expanded set of objects in Fig.~\ref{fig:robot_setup}(f)).
The skills were further expanded while we conducted the ablations to include instructions added in the last row of Table~\ref{table:all_skills}, which were used for the experiments described in Sec.~\ref{sec:exp_long} and~\ref{sec:exp_limit}.
These additional skills focused on realistic, long-horizon instructions in an office kitchen. %
The entire process of adding tasks and data is described in the Appendix~\ref{sec:app_data}.
Since we do not make any assumptions about particular skills when adding new instructions, the system is easily extendable, and we can continuously provide more diverse data to improve its capabilities.

\section{Experiments}\label{sec:exp}

Our experiments seek to answer the following questions: 
\begin{enumerate}[topsep=0pt,itemsep=-1ex,partopsep=1ex,parsep=1ex]
    \item Can an \algname learn to perform a large number of instructions, as well as to generalize in zero shot to new tasks, objects and environments? (Section~\ref{sec:exp_perf})
    \item Can we push the resulting model even further by incorporating heterogeneous data sources, such as simulated data or data from different robots?  (Section~\ref{sec:exp_limit})
    \item How do various methods generalize to long-horizon robotic scenarios? (Section~\ref{sec:exp_long})
    \item How do generalization metrics change with varying amounts of data quantity and data diversity? (Section~\ref{sec:exp_data})
    \item What are the important and practical decisions in the design of the model and how do they affect performance and generalization? (Appendix Section~\ref{sec:app_model_ablations})
\end{enumerate}

Throughout this section we will compare to two baseline state of the art architectures, Gato~\citep{reed2022generalist} and BC-Z~\citep{jang2022bc}. 
Importantly both of these are trained on our data described in detail in Sec.~\ref{sec:data} (which is an important part of our system) since the original models in these publications would not exhibit generalization properties required for our evaluation tasks.
Gato is, similarly to \algname, based on a Transformer architecture, but varies from \algname in multiple aspects. First, it computes image tokens without the notion of language and each image token embedding is computed separately for each image patch, as opposed to early language fusion and global image embedding in our model. Second, it does not use a pre-trained text embedding to encode the language string. It also does not include inference time considerations that are necessary for real robots as discussed in Sec.~\ref{sec:model} such as TokenLearner and the removal of auto-regressive actions. In order to run Gato on real robots at a high enough frequency, we also limit the size of the model compared to the original publication, which was 1.2B parameters (resulting in on robot inference time of $1.9$s), to be of similar size to \algname (37M parameters for Gato vs. 35M for \algname).
BC-Z is based on a ResNet architecture, and was used in SayCan~\citep{ahn2022can}. BC-Z differs from \algname in that it is a feedforward model that does not use previous timesteps, and it uses continuous actions rather than
discrete action tokens. In addition to the original BC-Z model size, we also compare our method to a larger version of BC-Z that has a similar number of parameters to \algname and refer to it as BC-Z XL.
We study and analyze how each of these design decisions changes performance in Appendix Sections~\ref{sec:app_model_ablations}~and~\ref{sec:app_analysis}.

We evaluate the success rate in experiments to measure performance on training instructions, generalization to unseen instructions, robustness to backgrounds and distractors, and performance in long-horizon scenarios, as detailed below.
Throughout this section, we evaluate our approach and baselines with over 3000 real-world trials, making one of the largest scale
evaluation of a robot learning system to-date.

\subsection{Experimental Setup}
\label{sec:exp_setup}
As mentioned in Section~\ref{sec:system}, we evaluate \algname with a set of mobile manipulators from Everyday Robots in three environments: two real office kitchens and a training environment modelled off these real kitchens. 
The training environment, shown in Fig.~\ref{fig:robot_setup} (a), consists of partial counters while the two real environments, shown in Fig.~\ref{fig:robot_setup} (b, c), have similar counter tops to the training environment, but vary in lighting, background, and full kitchen geometry (e.g., there may be a cabinet instead of a drawer or a sink may be visible).
The policies are evaluated for performance on training tasks as well as generalization to new tasks, robustness to unseen environments, and performance when chained together for long-horizon tasks, as detailed below.

\textbf{Seen task performance.}
To evaluate performance on seen instructions, we evaluate performance on instructions sampled from the training set. Note, however, that this evaluation still involves varying the placement of objects and other factors of the setup (e.g., time of day, robot position), requiring the skills to generalize to realistic variability in the environment.
In all, we test over 200 tasks in this evaluation: 36 for picking objects, 35 for knocking objects, 35 for placing things upright, 48 for moving objects, 18 for opening and closing various drawers, and 36 for picking out of and placing objects into drawers.

\textbf{Unseen tasks generalization.}
To evaluate generalization to unseen tasks, we test 21 novel, unseen instructions.
These instructions are distributed across skills and objects.
This ensures that at least some instances of each object and skill were present in the training set but they will be combined in novel ways. For example, if ``pick up the apple'' is held out, then there are other training instructions that include the apple. The list of all unseen instructions can be found in the Appendix~\ref{sec:app_exp}.

\begin{figure}[h]
     \centering
     \includegraphics[width=0.7\textwidth]{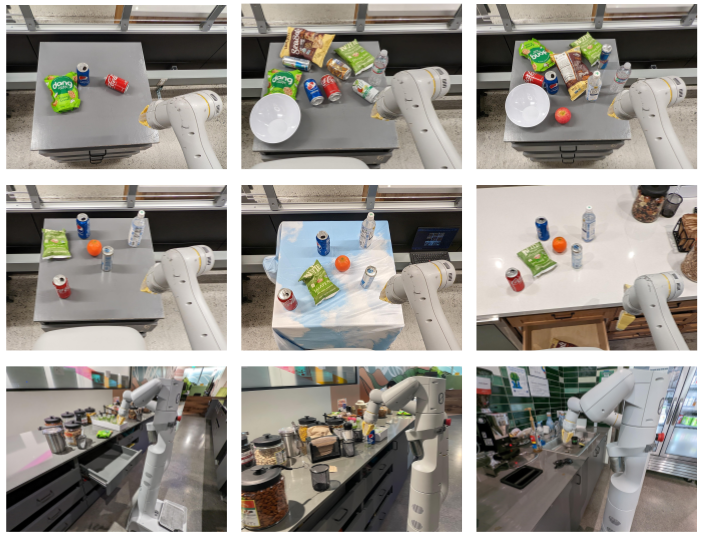}
     \caption{Evaluation scenarios for distractors (first row), from left to right: easy (0-5 distractors), medium (9 distractors), hard (9 distractors and occluded object); background (second row), from left to right: original environment, patterned table cloth, new kitchen; and realistic scenarios in the real kitchen (third row), generalization levels from left to right: $L1$, $L2$ and $L3$.}
     \label{fig:robustness}
\end{figure}

\textbf{Robustness.}
To evaluate robustness, we perform 30 real-world tasks for distractor robustness and 22 tasks for background robustness. 
The background robustness was tested by evaluating in new kitchens (which have different lighting and background visuals) and with different counter surfaces (e.g., a patterned table cloth). Example configurations of the robustness evaluation scenarios are depicted in Fig.~\ref{fig:robustness}.

\textbf{Long-horizon scenarios.}
We also evaluate generalization to more realistic long-horizon scenarios, which each require executing a sequence of skills. 
The goal of this evaluation is to combine multiple generalization axes such as new tasks, objects, environments and test the overall generalization capabilities in realistic settings. 
These evaluations consist of 15 long-horizon instructions in two real kitchens, which require executing sequences of skills consisting of $\sim10$ distinct steps, with each step of roughly comparable scope as the training instructions. These steps are obtained automatically from higher level instructions, such as ``how would you throw away all the items on the table?'' by using the SayCan system~\citep{ahn2022can}, as described in detail in Section~\ref{sec:exp_long} and Appendix~\ref{sec:app_exp_long}.

\subsection{Can \algname learn to perform a large number of instructions, and to generalize to new tasks, objects and environments?}
\label{sec:exp_perf}

To answer our first question, we analyze the overall performance, generalization, and robustness capabilities of \algname compared to previously proposed models. Specifically, we compare to the model architectures used by Gato~\citep{reed2022generalist} and BC-Z~\citep{jang2022bc}, as well as a larger version of BC-Z, which we refer to as BC-Z XL. Note, however, that all models are trained on the same data as \algname, and the evaluation only compares the model architectures, not the task sets, datasets, or overall robotic systems. 
The capabilities of \algname are determined to a large extent by the dataset and task set, which we believe improves significantly over prior works (e.g. BC-Z uses 100 tasks and the original Gato model trains a stacking task with various shapes), and thus this comparison should be viewed as rather favorable to the prior models, which also benefit from the large and diverse dataset and task set that we collected.

The results are shown in Table~\ref{table:main_baselines}.
Across each category, we find that \algname outperforms the prior models significantly.
On seen tasks, \algname is able to perform 97\% of the more than 200 instructions successfully, which is 25\% more than BC-Z and 32\% more than Gato.
On unseen tasks, \algname shows it is capable of generalizing to novel instructions, performing 76\% of the never-before-seen instructions, 24\% more than the next best baseline. 
While such generalization to novel instructions is made possible due to natural language conditioning of the policy, as the policy is able to understand new combinations of previously seen concepts, all of the baselines are also conditioned on natural language and in principle enjoy the same benefits.
We further ablate different components of \algname in the next section to better understand what aspects of our method contribute the most to this difference.
On distractors and backgrounds, we find that \algname is quite robust, successfully executing 83\% of the distractor robustness tasks and 59\% of the background robustness tasks (36\% and 18\% higher than the next best alternative, respectively).
Overall, we find that \algname has high general performance, while exhibiting impressive degrees of generalization and robustness.
We show example trajectories of the \algname agent including instructions that cover different skills, environments and objects in Fig.~\ref{fig:trajectories}.
We also present additional trajectory examples for different generalization tests in the Appendix, which include backgrounds (Fig.~\ref{fig:background_eval_trajectories}), and distractors (Fig.~\ref{fig:distractors_eval_trajectories}).

\begin{table}
	\begin{minipage}{0.45\linewidth}
		\setlength\tabcolsep{2.0pt}
        \scriptsize
        \begin{tabular}{@{}lccccccc@{}}
        \toprule
        Model & Seen Tasks & Unseen Tasks & Distractors & Backgrounds \\
        \midrule
        Gato~\citep{reed2022generalist} & 65  & 52 & 43 & 35 \\
        BC-Z~\citep{jang2022bc} & 72 & 19 & 47 & 41 \\
        BC-Z XL & 56 & 43 & 23 & 35 \\
        \algname (ours) & \textbf{97} & \textbf{76} & \textbf{83} & \textbf{59}
        \\
        \bottomrule
        \end{tabular}
	\end{minipage}\hfill
	\begin{minipage}{0.42\linewidth}
		\centering
        \includegraphics[width=\linewidth]{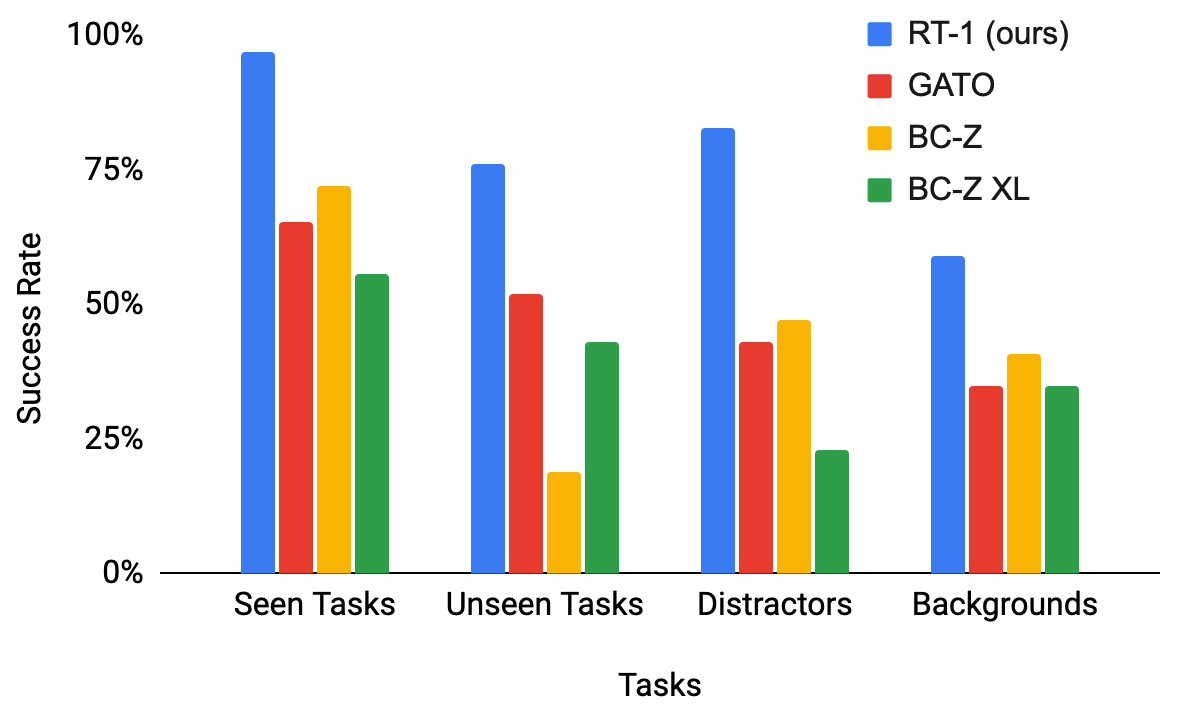}
        \caption{Overall performance of \algname and baselines across seen tasks, generalization to unseen tasks, and robustness to distractors and backgrounds.}
        \label{table:main_baselines}
	\end{minipage}
\end{table}

\begin{figure}[h!]
    \centering
    \includegraphics[width=0.8\linewidth]{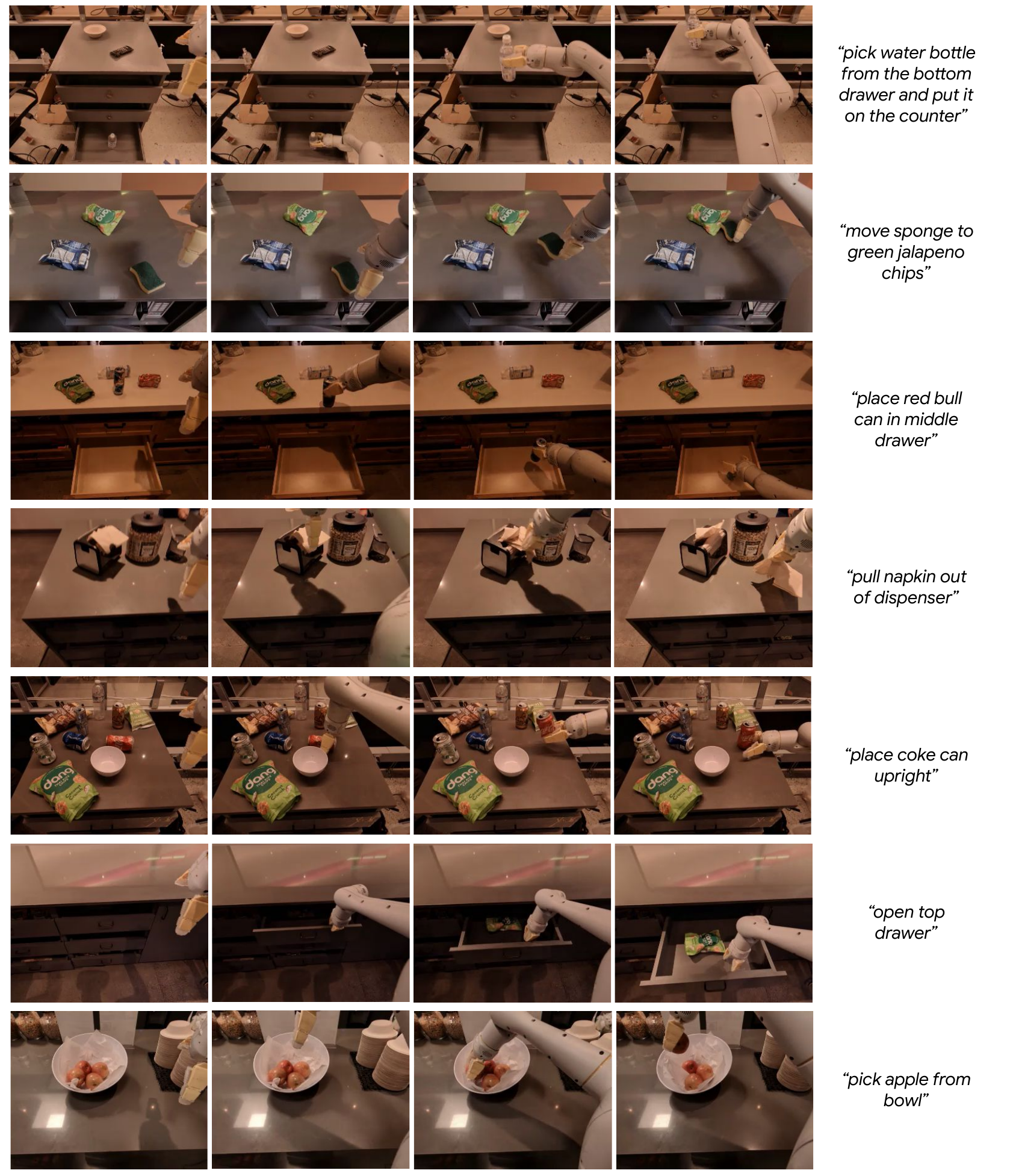}
    \caption{Example evaluation trajectories for \algname across various instructions.}
    \label{fig:trajectories}
\end{figure}

\textbf{Generalization to realistic instructions.}
Next, we test whether our method generalizes enough across all the different axes that we evaluated previously to be deployed in a real kitchen, which poses multiple distribution shifts all at once such as new tasks combinations, object distractors as well as a novel environment.

To evaluate our algorithm in realistic scenarios in a real kitchen, we construct task sequences to accomplish a number of realistic goals. The robot restocks several snacks in drawers, tidies up knocked over condiment bottles and closes drawers left open by humans, prepares a snack with an orange and a napkin and fetches lost sunglasses and an octopus toy from several places in the kitchen. The detailed instructions used in these scenarios are listed in the Appendix~\ref{sec:app_exp}.
The office kitchen involves a dramatic shift from the training environment and we categorize tasks across these scenarios with varying levels of generalization: $L1$ for generalization to the new counter-top layout and lighting conditions, $L2$ for additionally generalization to unseen distractor objects, $L3$ for additional generalization to drastically new task settings, new task objects or objects in unseen locations such as near a sink. The three levels that correspond to the three tasks of restocking, preparing a snack and fetching a lost object in the real kitchen are depicted in the last row of Fig.~\ref{fig:robustness}. Example trajectories for different levels are presented in the Appendix in Fig.~\ref{fig:generalization_eval_trajectories}.

We report the per-task success rate in these realistic scenarios along with the varying generalization levels in Table~\ref{table:levels} and find \algname to be the most robust on all levels. Gato generalizes fairly well at the first level but it performs significantly drops for the more difficult generalization scenarios. BC-Z and its XL equivalent perform fairly well at $L2$ level and better than Gato at $L3$ but they are still not at the generalization level of \algname.

\begin{table}
\begin{minipage}{0.5\linewidth}
  \setlength\tabcolsep{2.0pt}
  \scriptsize
  \begin{tabular}{@{}lcP{1cm}P{1cm}P{1cm}@{}}
  \toprule
  \multicolumn{1}{l}{} & & \multicolumn{3}{c}{Generalization Scenario Levels} \\
  \cmidrule(lr){3-5}
  Models & All &L1 & L2 & L3 \\
  \midrule
  Gato~\cite{reed2022generalist} & 30 & 63 & 25 & 0 \\
  BC-Z~\cite{jang2022bc} & 45 & 38 & 50 & \textbf{50}\\
  BC-Z XL & 55 & 63 & \textbf{75} & 38 \\
  \algname (ours) & \textbf{70} & \textbf{88} & \textbf{75} & \textbf{50}  \\
  \bottomrule
  \end{tabular}
  \end{minipage}
  \begin{minipage}{0.45\linewidth}
	\centering
    \scriptsize
    \includegraphics[width=1\linewidth]{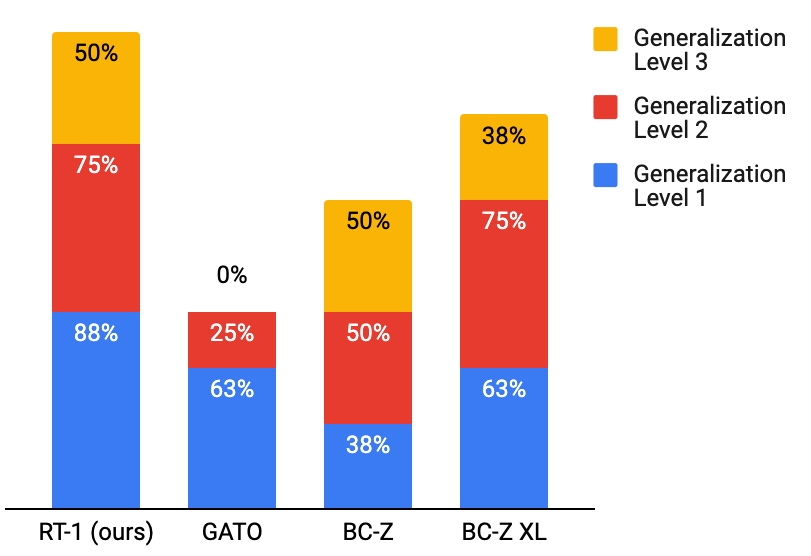}
\end{minipage}
\caption{\small{Realistic generalization scenarios: we compare model success rate in a realistic Google kitchen scenarios across three levels of generalization: $L1$ for generalization to the new counter-top layout and lighting conditions, $L2$ for additionally generalization to unseen distractor objects, $L3$ for additionally generalization to drastically new task settings, new task objects or in unseen locations like near a sink. }}
\label{table:levels}
\end{table}

\subsection{Can we push the resulting model further by incorporating heterogeneous data sources such as simulation or data from different robots?}\label{sec:exp_limit}

Next, we explore the limits of \algname for utilizing highly heterogeneous data.
We demonstrate how \algname can incorporate and learn from vastly different data sources and improve from such data without sacrificing its original-tasks performance across the varied tasks inherent in this data. 
To this end, we conduct two experiments: (1) \algname trained and tested on both real data and simulation data and (2) \algname trained across large datasets of  different tasks, originally collected by different robots. 
More information on each is provided in Appendix~\ref{sec:app_exp_limit}.

\textbf{Absorbing simulation data.}
Table~\ref{table:limit_sim} shows the ability of \algname, and baselines, to absorb both real and simulation data. 
To test this, we take all of the real demonstration data but we also provide additional simulation data that includes objects that the robot has never seen in the real world. 
Specifically, we specify different generalization scenarios: for \textit{seen skills with real objects} the training data has real data of that instruction (i.e., performance on seen tasks),
for \textit{seen skills with sim objects} the training data has sim data of that instruction (e.g. ``pick up a sim object'', which was present in sim), and for \textit{unseen skills with sim objects} the training data has sim data of that object but there are no examples of the instruction describing the skill with that object either in sim or in real (e.g., ``move a sim object to apple'', even though the robot has only practiced in picking that sim object and not moving it near other objects).
All evaluations are done in the real world but to limit the number of instructions evaluated, we focus on pick and move-to skills.

We find in Table~\ref{table:limit_sim} that for \algname, we do not lose performance adding simulation data compared to the Real Only dataset. 
We do however, see a significant increase in performance (from 23\% to 87\%) on objects and tasks seen only in simulation, to approximately the performance of the those in real, demonstrating an impressive degree of domain transfer.
We also see a significant increase in performance on unseen instructions from 7\%
to 33\%; impressive given the object in question has never been seen in real and the instruction never seen at all.
Overall, we find that \algname is able to efficiently absorb new data, even from a very different domain.

\begin{table}
	\begin{minipage}{0.45\linewidth}
		\setlength\tabcolsep{2.0pt}
  \scriptsize
  \begin{tabular}{@{}lcccc@{}}
  \toprule
  \multicolumn{2}{l}{} &  \quad Real Objects  \qquad & \multicolumn{2}{c}{ \quad Sim Objects (not seen in real)}\\
  \cmidrule(lr){3-3} \cmidrule(lr){4-5}
 & & Seen Skill   & \quad Seen Skill  &  Unseen Skill \\
Models & Training Data & w/ Objects & \quad w/ Objects & w/ Objects \\
  \midrule
  \algname & \textcolor{blue}{Real Only} & 92 & 23 & 7\\
  \algname & \textcolor{blue}{Real} + \color{ForestGreen}{Sim} & 90(-2) & \textbf{87(+64)} & \textbf{33(+26)}\\
  \bottomrule
  \end{tabular}
	\end{minipage}\hfill
	\begin{minipage}{0.42\linewidth}
		\centering
        \includegraphics[width=\linewidth]{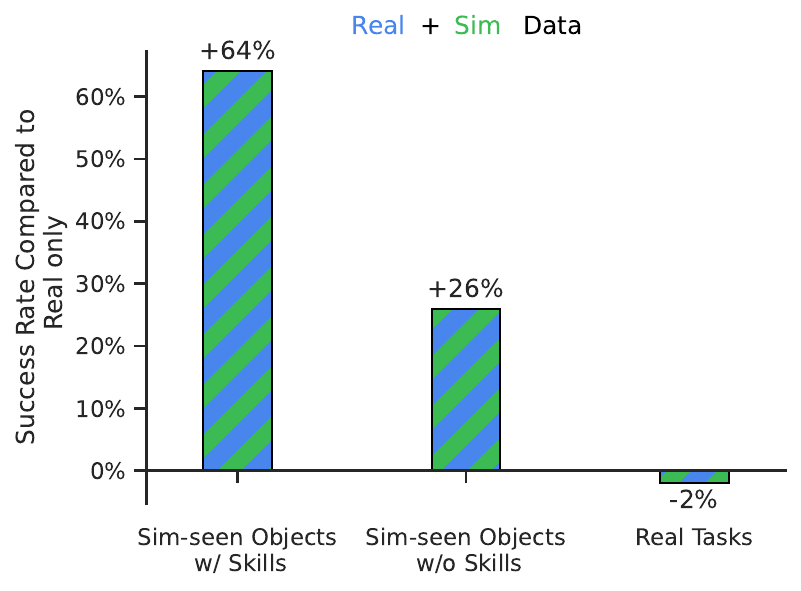}
        \caption{Experimental results for incorporating simulation data in \algname. Adding simulation data does not impact the performance on real objects, while significantly improving real performance on objects that were only introduced in simulation (+64\%). It also improves real-world generalization on simulated objects used with skills seen only in the real world (+26\%), e.g. ``move X to Y'' where X only appeared in simulated ``pick X'' task.}
  \label{table:limit_sim}
	\end{minipage}
\end{table}

\begin{figure}[h!]
    \centering
    \includegraphics[width=0.8\linewidth]{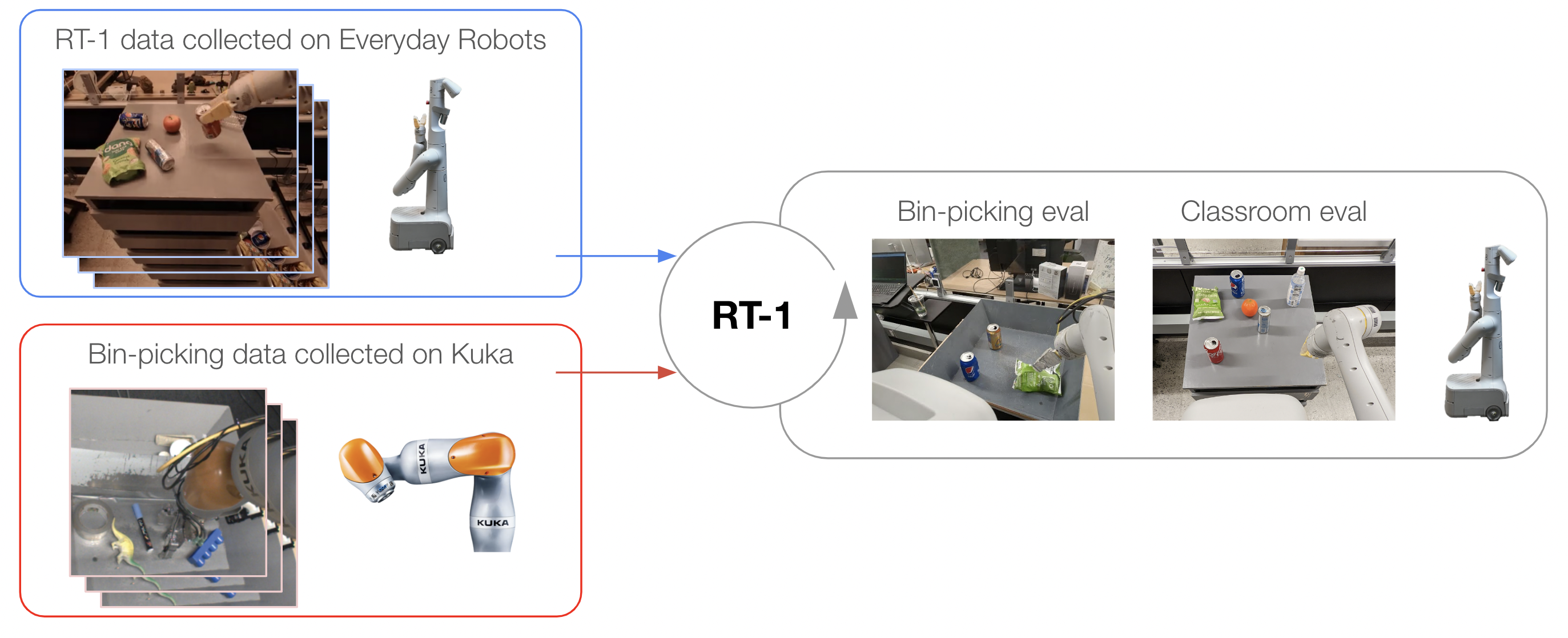}
    \caption{In Table~\ref{table:kuka}, \algname is trained with data from two robotics platforms and learns to generalize across them.}
    \label{fig:kuka_data_figure}
\end{figure}

\textbf{Absorbing data from different robots.}
To push the data absorption limits of \algname, we conduct an additional set of experiments where we combine two data sources that originate from different robots: Kuka IIWA as well as the Everyday Robots mobile manipulators used in the experiments so far. The Kuka data contains all the successful examples collected in QT-Opt~\citep{kalashnikov2018qtopt}, which corresponds to 209k episodes, where the robot was indiscriminately grasping objects in a bin (see an example of a Kuka episode in Table.~\ref{table:kuka}).
To test whether \algname can effectively absorb these two very different datasets, which we refer to as the standard ``Classroom eval'', as well as the performance on the newly constructed tasks that reflect the bin-picking setup present in the Kuka data, which we refer to as the ``Bin-picking eval'' (see Fig.~\ref{fig:kuka_data_figure}).  

We would like to emphasize the difficulty of this setting by noting the major differences between the datasets. Not only are the robots that collected the data different in appearance and action space, but also the environment they were deployed in has different appearance and dynamics. In addition the QT-Opt data presents a completely different action distribution -- it was collected by an RL agent as opposed to human demonstrations present in our dataset.

The results are presented in Table~\ref{table:kuka}. We observe that the model that mixes the \algname data and the Kuka data has only a minimal decrease in the original tasks' performance (i.e. Classroom eval), i.e. $2\%$. Even more importantly, in the Bin-picking eval, we observe that the model trained on multi-robot data performs at $39\%$ compared to the $22\%$ of the model that was trained only on the \algname data. This is a $17\%$ performance difference (almost 2x).
Additionally, \algname trained on Kuka bin-picking data and evaluated on the bin-picking tasks with the Everyday Robots (EDR) robot achieves 0\% performance, confirming that it is difficult to transfer a behavior from another robot morphology. However, mixing the data from both robots allows \algname to infer the correct actions of the EDR robot even when faced with the states observed by Kuka robots. This is achieved without explicit demonstrations of bin-picking on EDR robot and by taking advantage of past experiences collected by Kuka robots. 
These results indicate that \algname's absorption properties also include 
the ability to acquire new skills through observing other robots' experiences and present an exciting avenue of future work where we combine many more multi-robot datasets to enhance the robot capabilities.

\begin{table}
\begin{minipage}{0.55\linewidth}
  \setlength\tabcolsep{2.0pt}
  \scriptsize
  \begin{tabular}{@{}lccc@{}}
  \toprule
  Models & Training Data & Classroom eval  & Bin-picking eval \\
  \midrule
  \algname & \textcolor{red}{Kuka bin-picking data} + \textcolor{blue}{EDR data} & 90(-2) & \textbf{39(+17)}\\
  \midrule
  \algname & \textcolor{blue}{EDR only data} & 92 & 22\\
  \algname & \textcolor{red}{Kuka bin-picking only data} & 0 & 0\\
  \bottomrule
  \end{tabular}
  \end{minipage}
  \hfill
  \begin{minipage}{0.4\linewidth}
	\centering
    \scriptsize
    \includegraphics[width=1\linewidth]{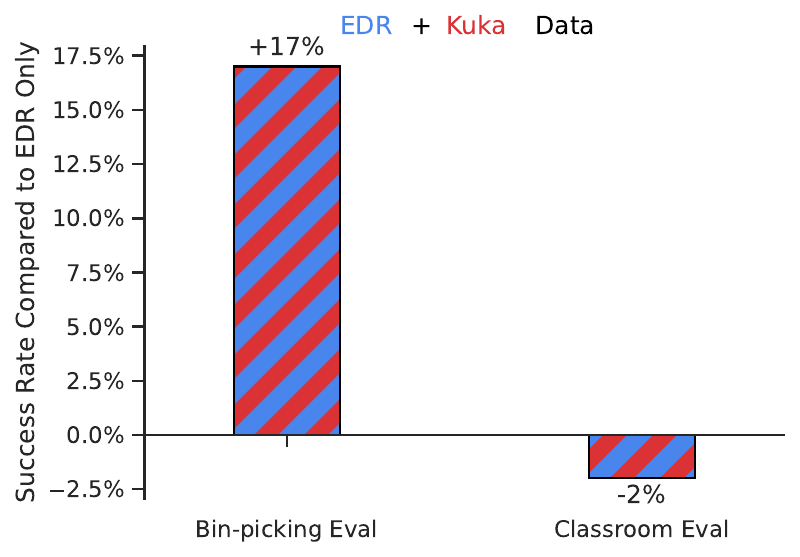}
\end{minipage}
\caption{Experimental results for mixing data from two different robots. Incorporating Kuka bin-picking data from QT-Opt~\citep{kalashnikov2018qtopt} in \algname minimally impacts the standard classroom evaluation performance and results in almost a 2x improvement in generalization to the Bin-picking evaluation (that is similar to the setup in the Kuka data) on the Everyday Robots manipulator. This demonstrates an effective transfer across two different robot morphologies.}
\label{table:kuka}
\end{table}

\subsection{How do various methods generalize long-horizon robotic scenarios?}
\label{sec:exp_long}
In the next set of experiments we evaluate whether our method generalizes enough to be used in long-horizon realistic kitchen settings.
To answer this question, we execute \algname and various baselines within the SayCan~\citep{ahn2022can} framework  in two different real kitchens.
Since SayCan combines many low-level instructions to perform high-level instructions, the number of possible high-level instructions increases combinatorially with skills, so the skill-breadth of \algname can be fully seen (for more details on the SayCan algorithm please refer to~\cite{ahn2022can}).
The success rate of long-horizon tasks also decreases exponentially with the length of the task, so high success rates in manipulation skills are particularly important. Furthermore, as mobile manipulation tasks require both navigation and manipulation, the policies ability to be robust to base position is crucial.
More detail is provided in Appendix~\ref{sec:app_exp_long}.

Table~\ref{table:saycan} shows our results (on instructions in Appendix Table~\ref{tab:long_horizon_all_instructions}). Except for original SayCan, all methods get 87\% as planning success rate, and RT-1 performs the best, with 67\% execution success rate in Kitchen1. Kitchen2 constitutes a much more challenging generalization scene, since the Robot Classroom training scenes are modeled after Kitchen1 (see the pictures of the kitchens in Fig.~\ref{fig:robot_setup}). Due to this generalization difficulty, SayCan with Gato is not able to finish any long horizon task, and SayCan with BC-Z is able to achieve a success rate of 13\%. The original SayCan paper did not evaluate performance in a new kitchen. 
Surprisingly, the manipulation performance does not see a visible drop from Kitchen1 to Kitchen2 for our method. In the supplementary video, we show that this enables us to operate unseen drawers in Kitchen2, and that we can use SayCan-RT1 to plan and execute ultra-long horizon tasks, with as many as 50 steps.

\begin{table}[h!]
\begin{center}
  \vspace{-1.0em}
  \setlength\tabcolsep{2.0pt}
  \scriptsize
  \begin{tabular}{@{}lcccccc@{}}
  \toprule
  \multicolumn{1}{l}{} & \multicolumn{2}{c}{SayCan tasks in Kitchen1} & \multicolumn{2}{c}{SayCan tasks in Kitchen2} &\\
  \cmidrule(lr){2-3} \cmidrule(lr){4-5} 
   & Planning & Execution &  Planning & Execution &  \\
  \midrule
 Original SayCan~\citep{ahn2022can}$^*$ & 73 & 47 & - & -\\
  SayCan w/ Gato~\citep{reed2022generalist} & 87 & 33 & 87 & 0\\
  SayCan w/ BC-Z~\citep{jang2022bc} & 87 & 53 & 87 & 13\\
  SayCan w/ \algname (ours) & 87 & \textbf{67} & 87 & \textbf{67}\\
  \bottomrule
  \end{tabular}
  \caption{SayCan style long horizon tasks in Kitchen1 and Kitchen2. (*Original SayCan eval uses a slightly different prompt so the planning success rate is lower.)}
  \label{table:saycan}
\end{center}
\end{table}

\subsection{How do generalization metrics change with varying amounts of data quantity and data diversity?}\label{sec:exp_data}

While previous works have shown the scaling abilities of Transformer-based models~\citep{lee2022multi,reed2022generalist, jiang2022vima} with the number of model parameters, in many robotics works the model size is often not the primary bottleneck, and the maximum size is limited by the latency requirement for running such models on real robots.
Instead, in this study we focus on ablating the influence of dataset size and diversity, as they play an important role in the traditionally data-limited robot learning field. 
Since data collection is particularly expensive for real robots, it is important to quantify what kind of data our models need to achieve a certain performance and generalization. 
Thus, our last question focuses on the scaling properties of \algname with different data properties.

\begin{table}[h]
	\begin{minipage}{0.7\linewidth}
	\centering
	\setlength\tabcolsep{2.0pt}
  \scriptsize
  \begin{tabular}{@{}lccccccccc@{}}
  \toprule
  \multicolumn{5}{l}{} & \multicolumn{4}{c}{Generalization} \\
  \cmidrule(lr){6-9}
  Models & \% Tasks & \% Data \quad\quad & Seen Tasks & \qquad & All & Unseen Tasks &  Distractors & Backgrounds \\
  \midrule
  Smaller Data \\
  \algname (ours) & 100 & 100 & 97 & \qquad & 73 & 76 & 83 & 59\\
  \algname & 100 & 51  & 71 & \qquad & 50 & 52 & 39 & 59\\
  \algname & 100 & 37  & 55 & \qquad & 46 & 57 & 35 & 47\\
  \algname & 100 & 22  & 59 & \qquad & 29 & 14 & 31 & 41\\
  \midrule
  Narrower Data\\
  \algname (ours) & 100 & 100 & 97 & \qquad & 73 & 76 & 83 & 59\\
  \algname & 75 & 97  & 86 & \qquad & 54 & 67 & 42 & 53\\
  \bottomrule
  \end{tabular}
	\end{minipage}\hfill
	\begin{minipage}{0.7\linewidth}
		\centering
		\vspace{0.5cm}
    \includegraphics[width=\linewidth]{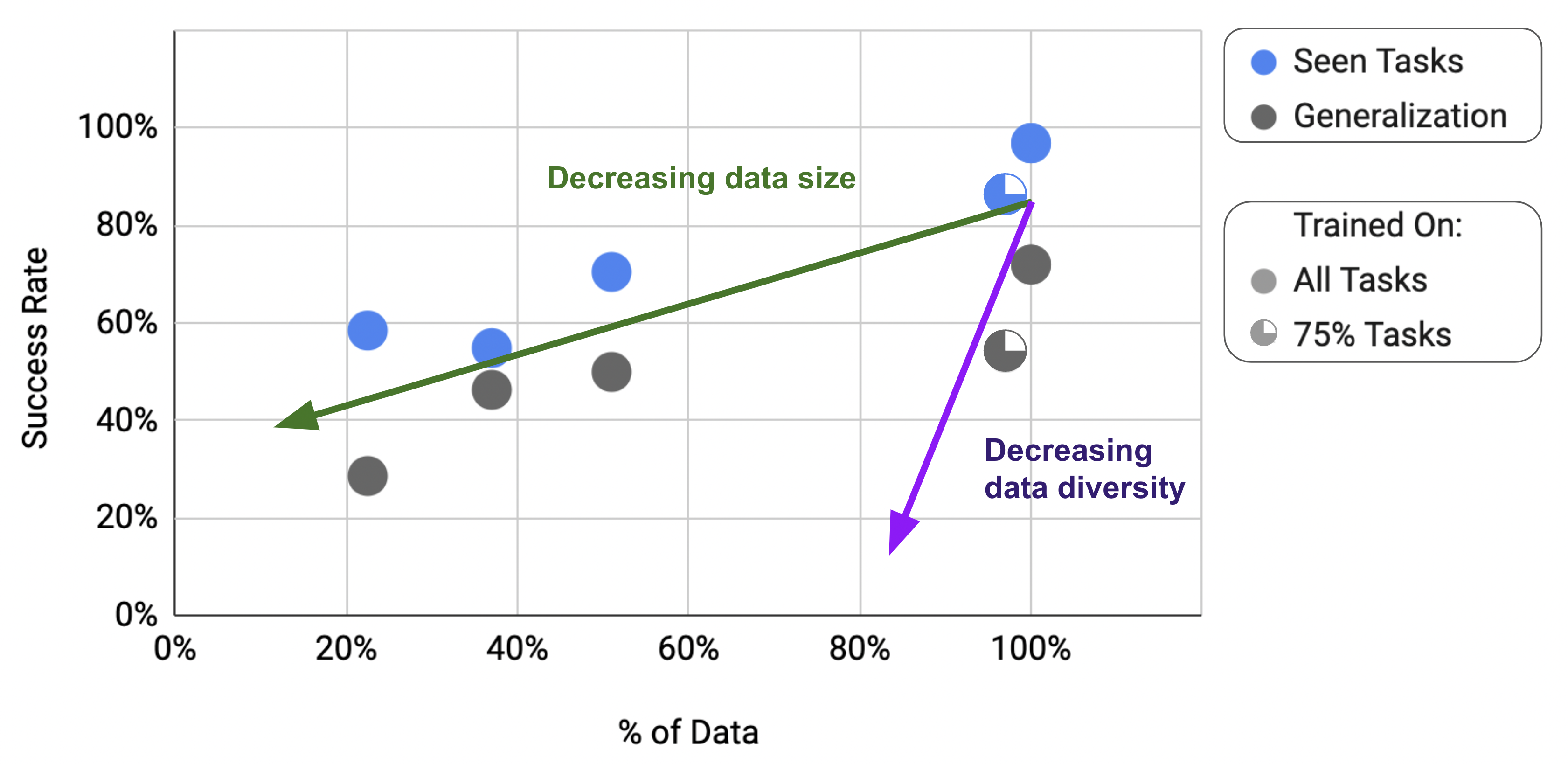}
    \caption{Various data ablations of \algname across seen tasks, generalization to unseen tasks, and robustness to distractors and backgrounds. Data diversity has a higher impact on the performance and generalization than data quantity.}
    \label{table:data_ablation}
	\end{minipage}
\end{table}

In Table~\ref{table:data_ablation} we show the performance, generalization, and robustness of \algname as we decrease the dataset size (\% data) and the dataset diversity (\% tasks). 
To separate the axes of dataset size and diversity, we create smaller datasets with the same task diversity by removing data from the tasks with the largest data, capping the number of examples per task at 200 (resulting in 51\% of the data), 100 (37\% of the data), and 50 (22.5\% of the data). 
To create a narrow dataset, we remove the tasks with the least data, thus keeping 97\% of the overall data but only 75\% of the tasks.
As we decrease dataset size, we see a general trend of decreasing performance and a steeper trend of decreasing generalization.
As we make the dataset more narrow, we see much steeper performance reductions, particularly in terms of generalization.
In fact, removing 25\% of the tasks while keeping 97\% of the data achieves an equivalent generalization performance to reducing the dataset size by as much as 49\%.
Our key takeaway is thus that data diversity is more essential than data quantity.

\section{Conclusions, Limitations and Future Work}
\label{sec:conclusions}

We presented \algfullname, \algname, a robot learning method that can effectively absorb large amounts of data and scales with data quantity and diversity. We trained \algname on a large dataset of demonstrations containing over 130k episodes collected over the course of 17 months with 13 robots.
In our broad set of experiments, we demonstrated that our method that can perform over 700 instructions at 97\% success rate and effectively generalize to new tasks, objects and environments better than previously published baselines. 
We also demonstrated that \algname can successfully absorb heterogeneous data from simulation and other robot morphologies without sacrificing original-tasks performance and while improving generalization to new scenarios.
Lastly, we showed how this level of performance and generalization allowed us to execute very long-horizon tasks in the SayCan~\citep{ahn2022can} framework, with as many as 50 steps.

While \algname presents a promising step towards large-scale robot learning with an data-absorbent model, it comes with a number of limitations. First, it is an imitation learning method, which inherits the challenges of that class of approaches such as the fact that it may not be able to surpass the performance of the demonstrators. Second, the generalization to new instructions is limited to the combinations of previously seen concepts and \algname is not yet able to generalize to a completely new motion that has not been seen before. Lastly, our method is presented on a large but not very dexterous set of manipulation tasks. We plan to continue extending the set of instructions that \algname enables and generalizes to to address this challenge.

As we explore future directions for this work, we hope to scale the number of robot skills faster by developing methods that allow non-experts to train the robot via directed data collection and model prompting. While the current version of \algname is fairly robust especially to distractor objects, its robustness to backgrounds and environments could be further improved by greatly increasing the environment diversity. We also hope to improve the reaction speeds and context retention of \algname through scalable attention and memory. 

To allow the research community to build on top of this work, we have open-sourced the code for \algname\footnote{\url{http://github.com/google-research/robotics_transformer}}, which we hope will provide researchers with a valuable resource for future research for scaling up robot learning.

\subsubsection*{Acknowledgments}

We would like to acknowledge Aleksandra Faust, Andy Christiansen, Chuyuan Fu, Daniel Kappler, David Rendleman, Eric Jang, Jessica Gomez, Jessica Lin, Jie Tan, Josh Weaver, Justin Boyd, Krzysztof Choromanski, Matthew Bennice, Mengyuan Yan, Mrinal Kalakrishnan, Nik Stewart, Paul Wohlhart, Peter Pastor, Pierre Sermanet, Wenlong Lu, Zhen Yu Song, Zhuo Xu, and the greater teams at Robotics at Google and Everyday Robots for their feedback and contributions.

\bibliography{references}
\bibliographystyle{iclr2023_conference}

\newpage
\appendix
\section*{Appendix}
\label{sec:app}

\section{Author Contributions}
\label{sec:app_contributions}

\begin{itemize}
    \item \textbf{Evaluations (ablations, designing procedures, implementations, and running ablations)}: Yevgen Chebotar, Keerthana Gopalakrishnan, Karol Hausman, Julian Ibarz, Brian Ichter, Alex Irpan, Isabel Leal, Kuang-Huei Lee, Yao Lu, Ofir Nachum, Kanishka Rao, Sumedh Sontakke, Austin Stone, Quan Vuong, Fei Xia, Ted Xiao, and Tianhe Yu.
    \item \textbf{Network Architecture (tokenizer, training, inference)}: Yevgen Chebotar, Keerthana Gopalakrishnan, Julian Ibarz, Alex Irpan, Kuang-Huei Lee, Yao Lu, Karl Pertsch, Kanishka Rao, Michael Ryoo, Sumedh Sontakke, Austin Stone, and Quan Vuong.
    \item \textbf{Developed Infrastructure (data, training, collect, simulation, evaluations, storage, and operations)}: Anthony Brohan, Keerthana Gopalakrishnan, Karol Hausman, Alex Herzog, Jasmine Hsu, Alex Irpan, Nikhil Joshi, Ryan Julian, Dmitry Kalashnikov, Yuheng Kuang, Isabel Leal, Yao Lu, Fei Xia, Ted Xiao, Peng Xu, Sichun Xu, and Tianhe Yu.
    \item \textbf{Leadership (managed or advised on the project)}: Chelsea Finn, Karol Hausman, Julian Ibarz, Sally Jesmonth, Sergey Levine, Yao Lu, Igor Mordatch, Carolina Parada, Kanishka Rao, Pannag Sanketi, Vincent Vanhoucke. 
    \item \textbf{Paper (figures, vizualizations, writing)}: Keerthana Gopalakrishnan, Karol Hausman, Brian Ichter, Sergey Levine, Ofir Nachum, Karl Pertsch, Kanishka Rao, Austin Stone, Fei Xia, and Ted Xiao.
    \item \textbf{Data collection and evaluations}: Noah Brown, Justice Carbajal, Joseph Dabis, Tomas Jackson, Utsav Malla, Deeksha Manjunath, Jodily Peralta, Emily Perez, Jornell Quiambao, Grecia Salazar, Kevin Sayed, Jaspiar Singh, Clayton Tan, Huong Tran, Steve Vega, and Brianna Zitkovich.
\end{itemize}

\section{Model Card}
\label{sec:app_model_card}
\begin{figure*}
\raggedright
\begin{framed}
\begin{center}
\large{\textbf{Model Card for RT-1 (Robotics Transformer)}}
\end{center}
\textbf{Model Details}
\begin{itemize}[topsep=2pt,itemsep=-0.5ex,partopsep=1ex,parsep=1ex]
\item Developed by researchers at Robotics at Google and Everyday Robots, 2022, v1.
\item Transformer-based model, built upon a FiLM-conditioned EfficientNet~\citep{pmlr-v97-tan19a}, a TokenLearner~\citep{ryoo2021tokenlearner}, and a Transformer~\citep{vaswani2017attention}.
\item Trained with imitation learning with inputs of natural language tasks and images and output robot actions.
\end{itemize}
    
\textbf{Intended Use}
\begin{itemize}[topsep=2pt,itemsep=-0.5ex,partopsep=1ex,parsep=1ex]
\item Intended to be used for controlling an Everyday Robot for manipulation tasks.
\item Unclear suitability as a learned representation for different robotic embodiments, environments, or significantly varied downstream tasks.
\item Not suitable for interaction with humans.
\end{itemize}

\textbf{Factors}
\begin{itemize}[topsep=2pt,itemsep=-0.5ex,partopsep=1ex,parsep=1ex]
\item Factors include varying backgrounds, lighting, scenes, base position, and novel natural language tasks. Hardware factors include camera and robot embodiment.
\end{itemize}

\textbf{Metrics}
\begin{itemize}[topsep=2pt,itemsep=-0.5ex,partopsep=1ex,parsep=1ex]
    \item Evaluation metrics include \textbf{seen task performance}, \textbf{unseen task performance}, \textbf{robustness} to backgrounds and distractors, and performance in \textbf{long-horizon scenarios}. Each measures the success rate of the model performing natural language specified tasks with randomized objects and object locations and varying scenes. 
\end{itemize}

\textbf{Training Data}
\begin{itemize}[topsep=2pt,itemsep=-0.5ex,partopsep=1ex,parsep=1ex]
    \item Trained on 130k tele-operation demonstrations over 13 robots and 744 tasks. 
\end{itemize}

\begin{minipage}{0.31\textwidth}

\begin{center}
\setlength\tabcolsep{2.0pt}
\scriptsize
\begin{tabular}{p{3.5cm} p{0.75cm} p{4.5cm} p{4cm}}
\toprule
Skill & Count & Description & Example Instruction\\
\midrule

Pick \texttt{Object} & 130 & Lift the object off the surface & pick iced tea can \\

Move \texttt{Object} Near \texttt{Object} & 337 & Move the first object near the second & move pepsi can near rxbar blueberry\\

Place \texttt{Object} Upright & 8 & Place an elongated object upright & place water bottle upright\\

Knock \texttt{Object} Over & 8 & Knock an elongated object over & knock redbull can over\\

Open / Close \texttt{Drawer} & 6 & Open or close any of the cabinet drawers & open the top drawer \\

Place \texttt{Object} into \texttt{Receptacle} & 84 & Place an object into a receptacle & place brown chip bag into white bowl\\

Pick \texttt{Object} from \texttt{Receptacle} and Place on the Counter & 162 & Pick an object up from a location and then place it on the counter  & pick green jalapeno chip bag from paper bowl and place on counter \\
\midrule
Additional tasks & 9 & Skills trained for realistic, long instructions & pull napkin out of dispenser \\

\midrule
Total & 744 \\
\bottomrule
\end{tabular}
\end{center}
\end{minipage}
\vspace{.5em}

\textbf{Evaluation Data}
\begin{itemize}[topsep=2pt,itemsep=-0.5ex,partopsep=1ex,parsep=1ex]
    \item Evaluated on real-world randomized scenes and over 3000 total rollouts in the environment it was trained on as well as two new office kitchen environments.
\end{itemize}

{\bf Quantitative Analyses}
\begin{itemize}[topsep=2pt,itemsep=-0.5ex,partopsep=1ex,parsep=1ex]
    \item \algname shows high-performance and robustness and can learn from heterogenous data.
\end{itemize}

\begin{minipage}{0.31\textwidth}
\vspace{.5em}
\begin{subfigure}{\textwidth}
\vspace{.25em}
\centering
\hbox{\includegraphics[scale=0.3, trim={0 0 0 0}, clip]
{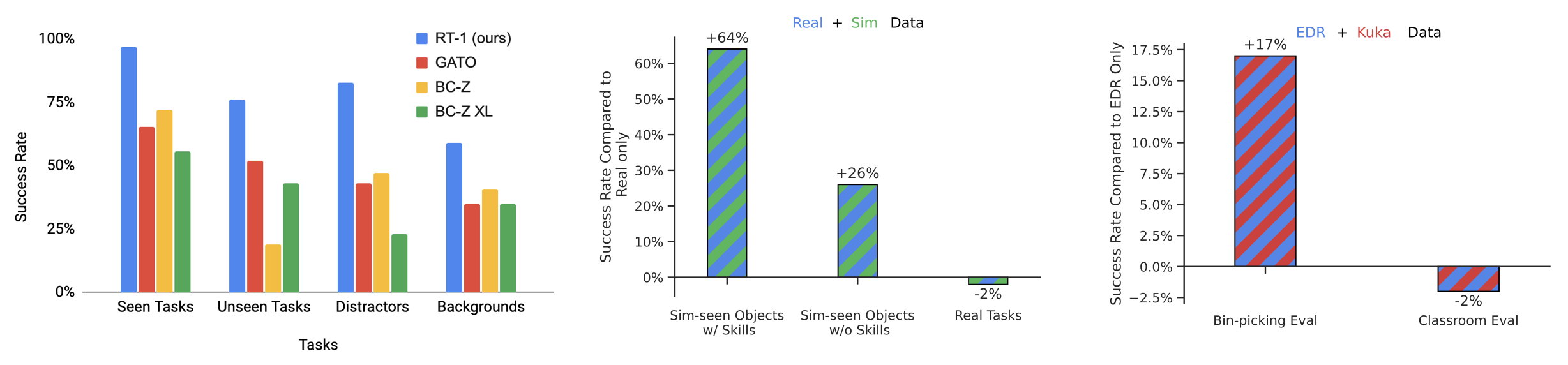}}
\label{fig:main_baselines_model_card}
\end{subfigure}
\end{minipage}

\textbf{Ethical Considerations}
\begin{itemize}[topsep=2pt,itemsep=-0.5ex,partopsep=1ex,parsep=1ex]
    \item Early research, model has not yet been evaluated for suitability to use outside of its current research setting.
\end{itemize}

\textbf{Caveats and Recommendations}
\begin{itemize}[topsep=2pt,itemsep=-0.5ex,partopsep=1ex,parsep=1ex]
    \item While the current model covers only a small portion of possible robotic manipulation tasks, it presents a recipe for scalable robotic learning and an architecture that shows favorable generalization and data absorption properties.
\end{itemize}

\end{framed}
\vspace{-1em}
\caption{Model Card for \algname.}
\label{fig:model_card}
\end{figure*}

We present the Model Card for \algname in Fig.~\ref{fig:model_card}.

\section{Model and Data}
\label{sec:app_model_data}

\subsection{Model inference}
\label{sec:app_inference}
In addition to the inference speed requirement, we need to ensure that our system outputs actions at a consistent frequency, avoiding jitter. To accomplish this, we introduce a fixed-time waiting mechanism that waits a certain amount of time ($280$ms, the max observed latency of all components) after the state, that was used to compute the next action, has been captured, but before applying the action, similarly to the procedure described by~\citet{xiao2020thinking}.

\subsection{Data collection at scale.}
\label{sec:app_collection}

Each of the robots autonomously approaches its station at the beginning of the episode and communicates to the operator the instruction that they should demonstrate to the robot. 
To ensure a balanced dataset as well as randomization of the scene, we created a software module responsible for sampling the instructions to be demonstrated as well as the randomization of the background configuration. 
Each of the robots tells the demonstrator how to randomize the scene and which instruction to demonstrate.

Demonstrations are collected with direct line-of-sight between operator and robot using 2 virtual reality remotes. We map remote controls onto our policy action space to preserve consistency of the transition-dynamics. 3D position and rotational displacements of the remote are mapped to 6d displacements of the robot tool. The x, y position of the joystick is mapped to a turning angle and driving distance of the mobile base. We compute and track trajectories to the target poses that we obtain from the joystick commands.

\subsection{Model Selection at Scale}
\label{sec:app_model_selection}
As robot learning systems become more capable and the number of instructions they can handle increases, evaluation of these models becomes difficult~\citep{kalashnikov2021mt,jang2022bc}.
This is an important consideration not only for evaluating different model classes and data distributions during the development process, but also for selecting the most performant model checkpoints for a particular training run.
While there have been a number of proposed solutions to this problem~\citep{dudik2011doubly, irpan2019off, hanna2017bootstrapping}, mostly known in the offline reinforcement learning literature as ``off-policy evaluation'', it still remains an open research challenge to evaluate multi-task robot learning systems at scale.

In this work, we propose leveraging simulation for ``real to sim'' transfer as a scalable tool that provides an approximate estimate of model performance during training across many real tasks.
We run policies trained from real data in a simulator to test the full rollout performance.
Note that all of our training data comes from the real world (except the experiment in Section~\ref{sec:exp_limit}), and the simulator is used only for model selection.
To accomplish this, we expand the simulation environment proposed by \citet{lee2022pi} to support 551 of the tasks described in Section~\ref{sec:data}.
For each of these tasks, we define a set of scene setup randomizations, robot pose randomizations, and success detection criteria.
To bridge the visual distribution shift between the real world and the simulation, we train a RetinaGAN~\citep{ho2020retinagan} model that transforms simulated images into realistic looking images.
Then, we deploy policies trained on real data directly into these simulation environments by applying RetinaGAN visual transformations at each timestep and measuring rollout simulated task success rates. 

While models trained only on real world data perform better in the real world than they do in simulation, we find that the simulation success rates of high-performing real world policies are higher than the simulation success rates of low-performing real world policies.
In other words, the \textit{ordering} of simulation policy success rates are informative for predicting the ordering of real world policy success rates.
We note that in this real-to-sim evaluation setting, we have a less strict requirement for simulation accuracy compared to sim-to-real settings; as long as simulation success rates are directionally correlated with real success rates, we can accept a moderate or even high gap between real and simulation success rates.

\begin{figure}[h]
     \centering
     \includegraphics[width=0.85\textwidth]{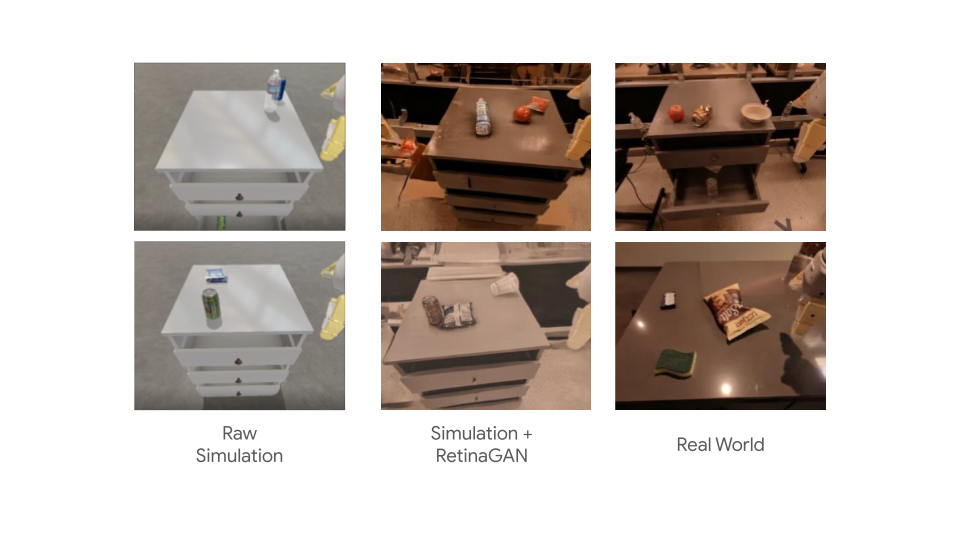}
     \caption{
{Example camera images showcasing raw simulation, simulation with RetinaGAN applied, and the real world.}
     }\label{fig:real2sim}
\end{figure}

We present example camera images from simulation as well as their RetinaGAN-based transformations in Fig.~\ref{fig:real2sim}.

\subsection{Data collection process}
\label{sec:app_data}

\begin{figure}[h!]
    \centering
    \includegraphics[width=0.8\linewidth]{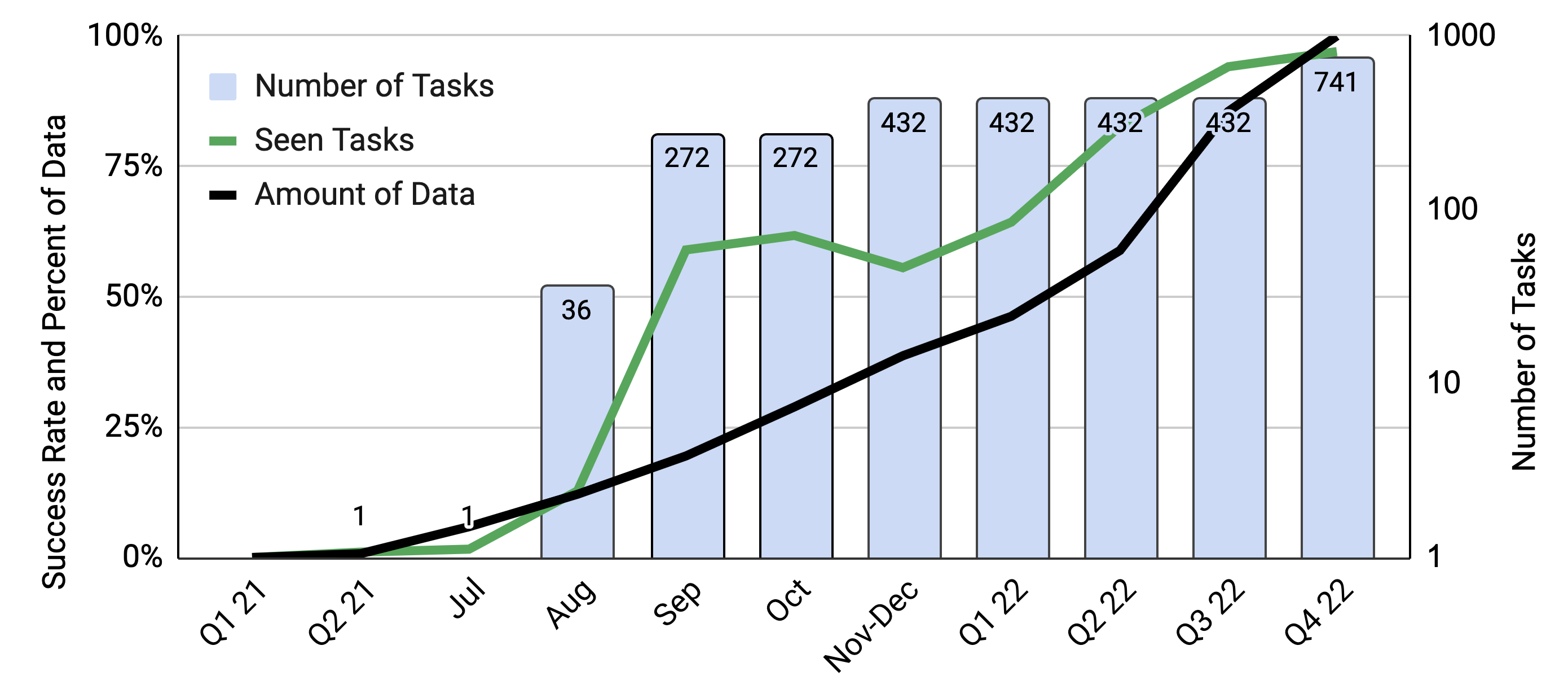}
    \caption{The growth of data, number of tasks, and seen instruction performance over time.}
    \label{fig:data_tasks_success}
\end{figure}

Figure~\ref{fig:data_tasks_success} shows the growth of data, number of tasks, and the success rate of the policy over time.
The number of tasks/instructions that our system is capable of grows over time as more data is collected. The same is true with the performance of seen tasks. One of the important aspects of the future work is develop techniques that allow us to grow the data as well as the robots performance and general capabilities at a faster rate.

\section{Experiments}
\label{sec:app_experiments}

\subsection{Evaluation Details}
\label{sec:app_exp}

In Section \ref{sec:exp_perf}, we study the zero-shot generalization capabilities of RT-1 to difficult scenarios not present in the training dataset. 
To fairly evaluate different ablations of RT-1 as well as baseline policies, we design standardized evaluation procedures that cover a range of incremental difficulty levels.

\textbf{Seen tasks.}
We evaluate on 744 tasks present in the training dataset.
The breakdown between 12 skills is shown in Table~\ref{table:all_skills}.
For all ``Seen'' evaluations, we use the same classroom setting used for data collection as described in Section~\ref{sec:data}.
For each policy, we report a single representative metric that takes a skill-weighted average across individual skill evaluations.

\textbf{Unseen tasks.}
We evaluate policy performance on 53 tasks that are held out during training. While the unseen instructions' specific combinations of skills and objects are not seen during training, other combinations of the same skills and objects are present in the training set.
We evaluate these unseen tasks in the same environment and the same randomization procedure as the \textbf{Seen tasks}.
A full list of these unseen tasks is shown in Table~\ref{tab:unseen_instructions}.

\textbf{Distractor robustness.}
We test three tasks (``pick coke can'', ``place coke can upright'', ``move coke can near green rice chip bag'') with incrementally more distractor objects added to the scene.
The easy setting includes 0, 2, or 5 distractor objects.
The medium setting includes 9 distractor objects, but the coke can is never obscured.
The hard setting includes 9 distractor objects, but the scene is more crowded and the coke can is partially occluded.
Both the medium are hard setting are more difficult than scenarios in the training dataset, which contained between 0 and 4 distractors.
Examples of these difficulty settings and policy evaluation rollouts are shown in Figure~\ref{fig:distractors_eval_trajectories}.

\begin{figure}[h!]
    \centering
    \includegraphics[width=\linewidth]{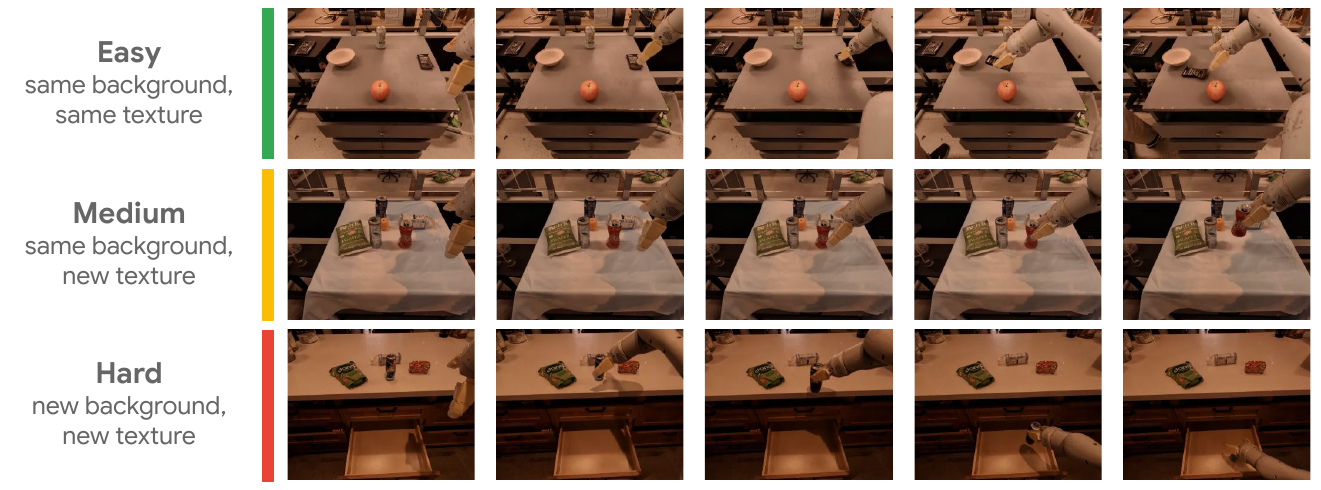}
    \caption{``Backgrounds'' evaluations focus on testing the performance of RT-1 on settings with different table textures and different backgrounds, such as those found in kitchens never trained on. These visual differences are quite pronounced, which in the most challenging case entails a new kitchen with different counter texture, different lighting conditions, different counter material, and a different background.}
    \label{fig:background_eval_trajectories}
\end{figure}

\textbf{Background robustness.}
We test six tasks (``pick coke can'', ``move blue chip bag near orange'', ``knock redbull can over'', ``pick green jalapeno chip bag'', ``move sponge near brown chip bag'',``place redbull can upright'') with incrementally more challenging backgrounds and counter textures.
In the easy setting, we utilize the same background environments and counter textures as the training dataset.
In the medium setting, we utilize the same background environment but add a patterned tablecloth to change the counter texture.
In the hard setting, we utilize a brand new kitchen environment with a new countertop; this changes the counter texture, drawer material and color, and background visuals.
Examples of these difficulty settings and policy evaluation rollouts are shown in Figure~\ref{fig:background_eval_trajectories}.

\begin{figure}[h!]
    \centering
    \includegraphics[width=\linewidth]{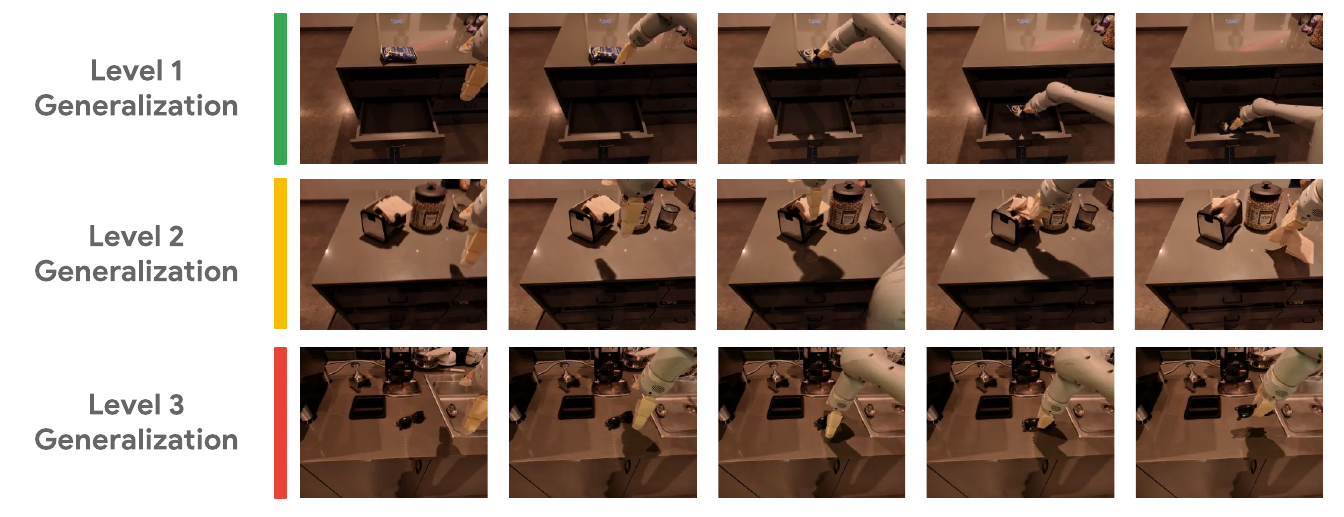}
    \caption{``Realistic instructions'' evaluations propose realistic scenarios multiple distribution shifts that incrementally increase in difficulty. $L1$ generalization introduces a new real office kitchen with new lighting conditions. $L2$ generalization additionally adds unseen distractor objects. Finally, $L3$ generalization includes new objects or objects in new locations, such as next to a sink.}
    \label{fig:generalization_eval_trajectories}
\end{figure}

\begin{figure}[h!]
    \centering
    \includegraphics[width=\linewidth]{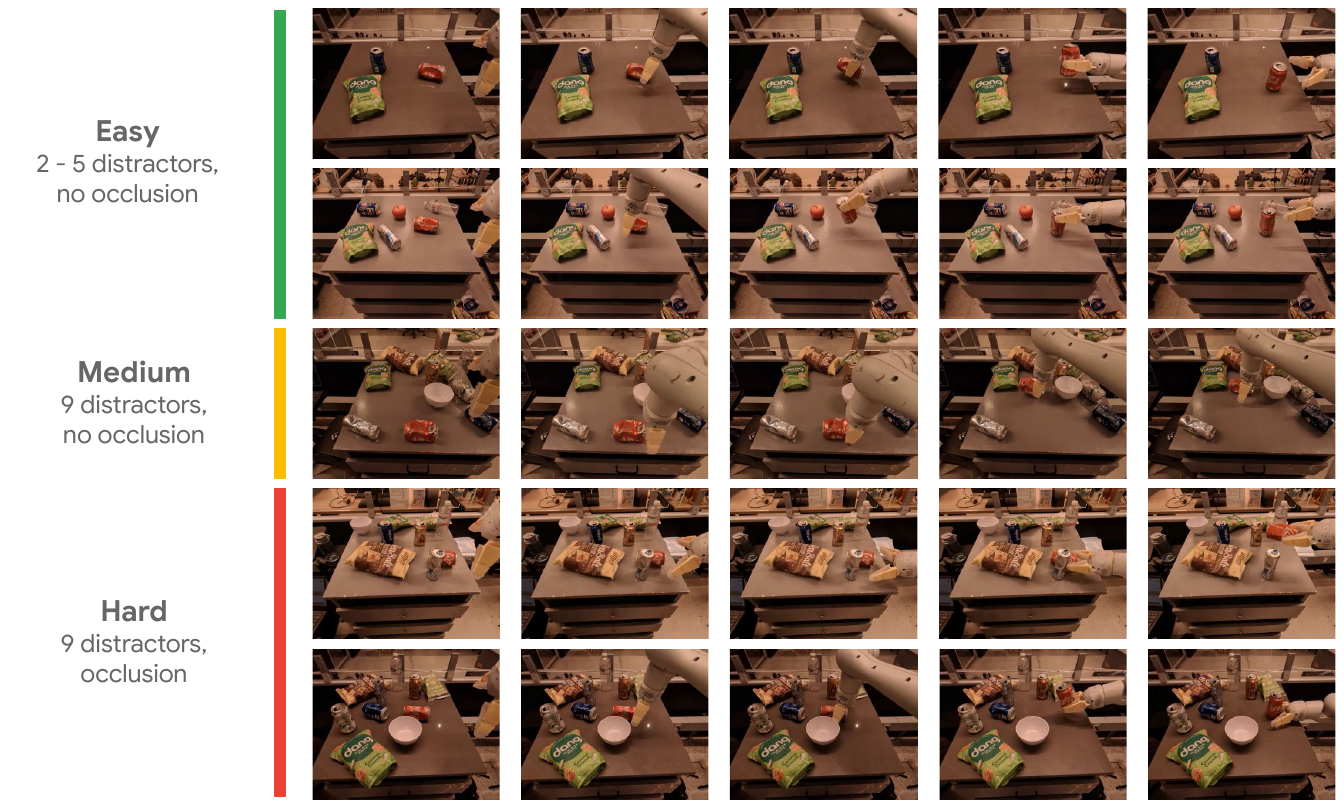}
    \caption{``Distractors'' evaluations focus on diversifying initial scene configurations well beyond the distributions contained in the training dataset, which contain between 2 and 4 distractor objects. In the most challenging scenarios, the scene is extremely cluttered and contains occlusions for the objects of interest.}
    \label{fig:distractors_eval_trajectories}
\end{figure}

\textbf{Realistic instructions.}
To study how RT-1 performs in more realistic scenarios, we propose an evaluation setting in a real office kitchen that is a dramatic shift from the original training classroom environment.
We propose a variety of skills that combine aspects of the previous zero-shot evaluations, including adding new distractors, including new backgrounds, and new combinations of objects with skills.
We refer to the easiest scenario as $L1$ generalization, which introduces a new countertop and lighting condition but keeps the skills and objects the same.
Next, $L2$ generalization additionally adds novel distractor objects such as kitchen jar containers.
Finally, $L3$ generalization adds new objects or new locations such as near a sink.
While some of these distribution shifts are tested in Section \ref{sec:exp_perf}, these realistic instructions aim to test multiple dimensions simultaneously.
Examples of these instructions are presented in Fig.~\ref{fig:generalization_eval_trajectories}.

\begin{table}
\small
\begin{center}

\begin{tabular}{|p{0.8\textwidth}|}
\hline
\textbf{Instruction} \\ \hline
pick coke can from top drawer and place on counter \\
pick green can from top drawer and place on counter \\
pick green rice chip bag from middle drawer and place on counter \\
pick redbull can from top drawer and place on counter \\
place 7up can into bottom drawer \\
place brown chip bag into top drawer \\
place green can into middle drawer \\
move 7up can near redbull can \\
move apple near green rice chip bag \\
move apple near paper bowl \\
move apple near redbull can \\
move blue chip bag near blue plastic bottle \\
move blue chip bag near pepsi can \\
move blue chip bag near sponge \\
move brown chip bag near apple \\
move brown chip bag near green rice chip bag \\
move brown chip bag near redbull can \\
move coke can near green jalapeno chip bag \\
move coke can near water bottle \\
move green can near 7up can \\
move green can near apple \\
move green can near coke can \\
move green jalapeno chip bag near blue chip bag \\
move green rice chip bag near orange \\
move green rice chip bag near orange can \\
move green rice chip bag near paper bowl \\
move orange can near brown chip bag \\
move pepsi can near orange can \\
move redbull can near coke can \\
move rxbar blueberry near blue plastic bottle \\
move rxbar blueberry near orange can \\
move rxbar chocolate near paper bowl \\
move rxbar chocolate near rxbar blueberry \\
move sponge near apple \\
move water bottle near 7up can \\
move water bottle near sponge \\
move white bowl near orange can \\
pick blue plastic bottle \\
pick green rice chip bag \\
pick orange \\
pick rxbar chocolate \\
pick sponge \\
place pepsi can upright \\
knock orange can over \\
pick blue plastic bottle from paper bowl and place on counter \\
pick brown chip bag from white bowl and place on counter \\
pick green can from paper bowl and place on counter \\
pick green jalapeno chip bag from white bowl and place on counter \\
pick orange can from white bowl and place on counter \\
pick redbull can from white bowl and place on counter \\
place blue plastic bottle into paper bowl \\
place coke can into paper bowl \\
place orange can into paper bowl \\
\hline
\end{tabular}
\caption{\textbf{List of Unseen Instructions in Sec.~\ref{sec:exp_perf}}. For the ``Unseen Tasks'' evaluation, we exclude a total of 53 tasks during training. While these exact instructions were not present in the training set, the objects and skills contained in these instructions were still present in the training set.}
\label{tab:unseen_instructions}
\end{center}
\end{table}

\subsection{Heterogeneous Data}
\label{sec:app_exp_limit}

We also explore the limits of \algname for utilizing highly heterogeneous data.
We demonstrate how \algname can incorporate and learn from vastly different data sources and improve from such data without sacrificing its original-tasks performance across the varied tasks inherent in this data. 
To this end, we conduct two experiments: (1) \algname trained and tested on both real data and simulation data and (2) \algname trained across large datasets of  different tasks, originally collected by different robots. 

\textbf{Absorbing simulation data.}
Table~\ref{table:app_limit_sim} shows the ability of \algname, and baselines, to absorb both real and simulation data. 
To test this, we take all of the real demonstration data but we also provide additional simulation data that includes objects that the robot has never seen in the real world. 
We add a set of sim objects and only show them on a subset of tasks, specifically the picking tasks, in simulation. To accomplish this, we run our real2sim method described in Sec.~\ref{sec:app_model_selection} to bootstrap a simulation policy from the real world policy that is then trained with multi-task RL~\citep{kalashnikov2021mt} with additional objects in simulation. 
From this process, we extract 518k successful trajectories of picking new objects and mix them with the real data that was used in the previous experiments. 
The goal of this experiment is to demonstrate that by expanding the dataset of simulation trajectories, we can benefit \algname's generalization capabilities without sacrificing the original training performance -- a desired property of an absorbent model. 

To evaluate the properties of this model, we specify different generalization scenarios: for \textit{seen skills with real objects} the training data has real data of that instruction (i.e., performance on seen tasks),
for \textit{seen skills with sim objects} the training data has sim data of that instruction (e.g. ``pick up a sim object'', which was present in sim), and for \textit{unseen skills with sim objects} the training data has sim data of that object but there are no examples of the instruction describing the skill with that object either in sim or in real (e.g., ``move a sim object to apple'', even though the robot has only practiced in picking that sim object and not moving it near other objects).
All evaluations are done in the real world but to limit the number of instructions evaluated, we focus on pick and move-to skills.

We find in Table~\ref{table:app_limit_sim} that for \algname, we do not lose performance adding simulation data compared to the Real Only dataset. 
We do however, see a significant increase in performance (from 23\% to 87\%) on objects and tasks seen only in simulation, to approximately the performance of the those in real, demonstrating an impressive degree of domain transfer.
We also see a significant increase in performance on unseen instructions from 7\%
to 33\%; impressive given the object in question has never been seen in real and the instruction never seen at all.
Overall, we find that \algname is able to efficiently ``sponge up'' new data, even from a very different domain.

\begin{table}
	\begin{minipage}{0.45\linewidth}
		\setlength\tabcolsep{2.0pt}
  \scriptsize
  \begin{tabular}{@{}lcccc@{}}
  \toprule
  \multicolumn{2}{l}{} & Real Objects  & \multicolumn{2}{c}{Sim Objects (not seen in real)}\\
  \cmidrule(lr){3-3} \cmidrule(lr){4-5}
 & & Seen Skill   & Seen Skill  &  Unseen Skill \\
Models & Training Data & w/ Objects & w/ Objects & w/ Objects \\
  \midrule
  \algname & Real Only & 92 & 23 & 7\\
  \algname & Real + Sim & 90 & \textbf{87} & \textbf{33}\\
  \bottomrule
  \end{tabular}
	\end{minipage}\hfill
	\begin{minipage}{0.42\linewidth}
		\centering
        \includegraphics[width=\linewidth]{sim_delta.pdf}
        \caption{Experimental results for incorporating simulation data in \algname. Adding simulation data does not impact the performance on real objects, while significantly improving real performance on objects that were only introduced in simulation.}
  \label{table:app_limit_sim}
	\end{minipage}
\end{table}

\textbf{Absorbing data from different robots.}
To push the data absorption limits of \algname, we conduct an additional set of experiments where we combine two data sources that originate from different robots: Kuka IIWA as well as the Everyday Robots mobile manipulators used in the experiments so far. The Kuka data contains all the successful examples collected in QT-Opt~\citep{kalashnikov2018qtopt}, which corresponds to 209k episodes, where the robot was indiscriminately grasping objects in a bin (see an example of a Kuka episode in Table.~\ref{table:app_kuka}).
Our goal in this experiment is to analyze whether the performance on the \algname tasks drops when adding the additional data and, more importantly, whether we can observe any transfer from data collected by a different robot morphology.

We would like to emphasize the difficulty of this setting by noting the major differences between the datasets. Not only are the robots that collected the data different in appearance and action space, but also the environment they were deployed in has different appearance and dynamics. In addition the QT-Opt data presents a completely different action distribution -- it was collected by an RL agent as opposed to human demonstrations present in our dataset.

To mix the Kuka data together with the \algname data, we first transform the original Kuka 4-DOF action space into the same action space as \algname, namely we set the roll and pitch to $0$, while keeping the yaw values that were present in the original Kuka data. In addition, we transform the binary \textit{gripper-close} command into a continuous gripper-closedness command that is present in the \algname data. We also need text instructions corresponding to the task performed and since the Kuka data does not contain the name of the object that was grasped, we relabel all the data to the ``pick anything'' instruction. With these modifications, we mix both datasets with the 2:1 (\algname data : Kuka data) ratio and train \algname to obtain the final model.

To test whether \algname can effectively absorb these two very different datasets, we evaluate the performance on the original \algname tasks (in this case, we also focus on ``pick'' and ``move to'' skills), which we refer to as the standard ``Classroom eval'', as well as the performance on the newly constructed tasks that reflect the bin-picking setup present in the Kuka data, which we refer to as the ``Bin-picking eval''.  
For the Bin-picking eval to be close to the original dataset, we put in the same looking bin for the objects as well as modify the robot to be similar to the Kuka manipulators by adding extra wires and coloring the gripper gray. For all of the evaluations we use the Everyday Robots robot with the picking commands and evaluate it based on 72 grasping trials.

The results are presented in Table~\ref{table:app_kuka}. We observe that the model that mixes the \algname data and the Kuka data has only a minimal decrease in the original tasks' performance (i.e. Classroom eval), i.e. $2\%$. Even more importantly, in the Bin-picking eval, we observe that the model trained on multi-robot data performs at $39\%$ compared to the $22\%$ of the model that was trained only on the \algname data. This is a $17\%$ performance difference (almost 2x).
Additionally, \algname trained on Kuka bin-picking data and evaluated on the bin-picking tasks with the Everyday Robots (EDR) robot achieves 0\% performance, confirming that it is difficult to transfer a behavior from another robot morphology. However, mixing the data from both robots allows \algname to infer the correct actions of the EDR robot even when faced with the states observed by Kuka robots. This is achieved without explicit demonstrations of bin-picking on EDR robot and by taking advantage of past experiences collected by Kuka robots. 
These results indicate that \algname's absorption properties also include 
the ability to acquire new skills through observing other robots' experiences and present an exciting avenue of future work where we combine many more multi-robot datasets to enhance the robot capabilities.

\begin{table}
\begin{minipage}{0.55\linewidth}
  \setlength\tabcolsep{2.0pt}
  \scriptsize
  \begin{tabular}{@{}lccc@{}}
  \toprule
  Models & Training Data & Classroom eval  & Bin-picking eval \\
  \midrule
  \algname & Kuka bin-picking data + EDR data & 90 & \textbf{39}\\
  \midrule
  \algname & EDR only data & 92 & 22\\
  \algname & Kuka bin-picking only data & 0 & 0\\
  \bottomrule
  \end{tabular}
  \end{minipage}
  \hfill
  \begin{minipage}{0.4\linewidth}
	\centering
    \scriptsize
    \includegraphics[width=1\linewidth]{kuka_delta.pdf}
\end{minipage}
\caption{Experimental results for mixing data from two different robots. Incorporating Kuka bin-picking data from QT-Opt~\citep{kalashnikov2018qtopt} in \algname minimally impacts the standard classroom evaluation performance and results in almost a 2x improvement in generalization to the Bin-picking evaluation (that is similar to the setup in the Kuka data) on the Everyday Robots manipulator. This demonstrates an effective transfer across two different robot morphologies.}
\label{table:app_kuka}
\end{table}

\subsection{Long-horizon Evaluation Details}
\label{sec:app_exp_long}

In addition to short-horizon individual skill evaluations shown in previous sections, we also evaluate how RT-1 performs in a long-horizon realistic kitchen setting that chains multiple manipulation and navigation skills to accomplish natural language instructions within the SayCan framework \citep{ahn2022can}.
A list of long-horizon instructions used for these evaluations is listed in  Table~\ref{tab:long_horizon_all_instructions}.

The success rate of long-horizon tasks decreases exponentially with the length of the task, so high success rates in manipulation skills are particularly important. Furthermore, as mobile manipulation tasks require both navigation and manipulation, the policies ability to be robust to base position is crucial.
Since SayCan combines many low-level instructions to perform high-level instructions, the number of possible high-level instructions increases combinatorially with instructions, so the skill-breadth of \algname can be fully seen.

SayCan works by grounding language models in robotic affordances and it leverages few-shot prompting to break down a long horizon task expressed in natural language to a sequence of low level skills. An example of long horizon task would be ``Bring me two different sodas", and one feasible plan would be ``1. find a coke, 2. pick up the coke, 3. bring it to you, 4. put down the coke, 5. find a pepsi, 6. pick up the pepsi, 7. bring it to you, 8. put down the pepsi, 9. done.'' To obtain the affordance function we use value functions trained with MT-OPT~\citep{kalashnikov2021mt}. For a detailed description of SayCan algorithm please refer to ~\citep{ahn2022can}.

Since the focus of this paper is acquisition of many generalizable skills, we focus our evaluation on one subset of tasks presented in ~\cite{ahn2022can}. It is the long-horizon family of tasks, involving 15 instructions, each instruction requires an average of 9.6 steps to complete, and involves an average of 2.4 manipulation skills per instruction. A full list of the instructions can be found in Table~\ref{tab:long_horizon_all_instructions}.

We compare against 3 baselines. 1) SayCan with BC-Z, which uses SayCan planning algorithm with BC-Z as manipulation policy, 2) SayCan with Gato, which uses SayCan planning algorithm with Gato as manipulation policy, 3) Originally reported SayCan results, which use SayCan planning algorithm with BC-Z, but since it uses a slightly different prompt, the planning success rate is lower. We reimplemented 3) in 1) for a fair comparison. 

As shown in Table~\ref{table:app_saycan}, except for original SayCan, all methods get 87\% as planning success rate, and RT-1 performs the best, with 67\% execution success rate in Kitchen1. Kitchen2 constitutes a much more challenging generalization scene, since the Robot Classroom training scenes are modeled after Kitchen1 (see the pictures of the kitchens in Fig.~\ref{fig:robot_setup}). Due to this generalization difficulty, SayCan with Gato is not able to finish any long horizon task, and SayCan with BC-Z is able to achieve a success rate of 13\%. The original SayCan paper did not evaluate performance in a new kitchen. 
Surprisingly, the manipulation performance does not see a visible drop from Kitchen1 to Kitchen2 for our method. In the supplementary video, we show that this enables us to operate unseen drawers in Kitchen2, and that we can use SayCan-RT1 to plan and execute ultra-long horizon tasks, with as many as 50 steps.

\begin{table}[h!]
\begin{center}
  \vspace{-1.0em}
  \setlength\tabcolsep{2.0pt}
  \scriptsize
  \begin{tabular}{@{}lcccccc@{}}
  \toprule
  \multicolumn{1}{l}{} & \multicolumn{2}{c}{SayCan tasks in Kitchen1} & \multicolumn{2}{c}{SayCan tasks in Kitchen2} &\\
  \cmidrule(lr){2-3} \cmidrule(lr){4-5} 
   & Planning & Execution &  Planning & Execution &  \\
  \midrule
 Original SayCan~\citep{ahn2022can}$^*$ & 73 & 47 & - & -\\
  SayCan w/ Gato~\citep{reed2022generalist} & 87 & 33 & 87 & 0\\
  SayCan w/ BC-Z~\citep{jang2022bc} & 87 & 53 & 87 & 13\\
  SayCan w/ \algname (ours) & 87 & \textbf{67} & 87 & \textbf{67}\\
  \bottomrule
  \end{tabular}
  \caption{SayCan style long horizon tasks in Kitchen1 and Kitchen2. (*Original SayCan eval uses a slightly different prompt so the planning success rate is lower.)}
  \label{table:app_saycan}
\end{center}
\end{table}

\begin{table}
\small
\begin{center}

\begin{tabular}{|p{0.8\textwidth}|}
\hline
\textbf{Instruction} \\ \hline
How would you put an energy bar and water bottle on the table  \\
How would you bring me a lime soda and a bag of chips  \\
Can you throw away the apple and bring me a coke \\
How would you bring me a 7up can and a tea?\\
How would throw away all the items on the table?  \\
How would you move an multigrain chips to the table and an apple to the far counter? \\
How would you move the lime soda, the sponge, and the water bottle to the table? \\
How would you bring me two sodas? \\
How would you move three cokes to the trash can? \\
How would you throw away two cokes?  \\
How would you bring me two different sodas? \\
How would you bring me an apple, a coke, and water bottle? \\
I spilled my coke on the table, how would you throw it away and then bring me something to help clean?  \\
I just worked out, can you bring me a drink and a snack to recover? \\
How would you bring me a fruit, a soda, and a bag of chips for lunch \\
\hline
\end{tabular}
\caption{\textbf{List of SayCan instructions evaluated in Sec.~\ref{sec:exp_long}}}
\label{tab:long_horizon_all_instructions}
\end{center}
\end{table}

\subsection{Model Ablations}
\textbf{What are the important and practical decisions in the design of the model and how do they affect performance and generalization?}
\label{sec:app_model_ablations}

To answer this question, we perform a set of ablations over different design decisions in \algname.
We aim to test a number of hypotheses that will help us disambiguate where the benefits of our method come from. 
Possible hypotheses about the source of improvement include: (i) the capacity and expressiveness of our model, which we verify by ablating the model size, trying other architectures (e.g., by removing the Transformer component); (ii) the particular action representation, which makes it easy to represent complex multi-modal action distributions, which we test by switching to continuous (normally distributed) actions, as well as by ablating the auto-regressive action representation;
(iii) the ImageNet pre-trained initialization of the components,
which we test by initializing the model's weights randomly; and (iv) access to the short history, which we test by excluding observation history.
More concretely, we ablate our model by (1) decreasing the model size (from 35M to 21M parameters), (2) removing the Transformer architecture (using a pre-trained EfficientNet instead), (3) using a continuous instead of discrete action space (using an MSE loss and multivariate normal output), (4) auto-regressively conditioning on actions,
(5) removing ImageNet pre-training of the FiLM EfficientNet, and (6) removing history (reducing the sequence of six images as input to a single image). 
For each ablation we compare on the axes of performance on seen tasks, performance on unseen tasks, as well as inference speed and robustness to distractors and backgrounds (with a more detailed description of each category in Section~\ref{sec:exp_setup} and Appendix~\ref{sec:app_exp}).

Table~\ref{table:model_ablation} shows the results of each ablation and the delta performance compared to the full \algname.
\algname achieves impressive performance on tasks and new environments, and particularly outperforms baselines on the most challenging robustness problems.
We also find that each design decision is important, though at varying levels. 
We first evaluate a model that replaces the per-dimension discretized action representation in our model with a more standard continuous Gaussian distribution. We observe a significant decline in performance from this modification. 
The per-dimension discretization allows our model to represent complex multi-modal distributions, while the Gaussian distribution captures only a single mode. These results suggest that this standard and popular choice is highly suboptimal with the more complex and diverse demonstration data used by our system.
ImageNet pre-training is particularly important for model generalization and robustness, decreasing the unseen task performance rate by 33\%, as a result of the large and diverse visuals of the ImageNet dataset.
Adding history has an impact primarily on generalization to distractors, while removing the Transformer
component has a uniform but small negative impact across the seen tasks, unseen tasks and distractors.
In order to keep the ImageNet pre-training while reducing the model size, we reduce the number of parameters only by 40\% (from 31M to 25M). 
Resulting performance drops across training and generalization tasks but not as much as in other ablations.
Finally, autoregressively conditioning on actions, as used in~\citep{reed2022generalist, chen2021decision, lee2022multi}, did not benefit performance and slowed inference by more than 2x.

As described in Sec.~\ref{sec:model}, in order to run large Transformer models on real robots, we require a model that supports fast inference for real-time operation. 
Note that in order to achieve our target control rate of $3$Hz (described in Sec.~\ref{sec:model}), we also need to consider other sources of latency in the pipeline, such as the camera latency and communication overhead. However, these factors will be constant for all the models, and therefore we focus our evaluation on just the network inference time.
The last column of Table~\ref{table:model_ablation} shows the inference speed of all the models.
\algname is almost an order of magnitude faster than Gato with a similar number of parameters, but it is also considerably slower than a ResNet-based BC-Z. 
In terms of the different ablations of our model, we observe that the biggest slow-down is caused by including auto-regressive actions ($\sim$2x slow-down), and since this does not significantly influence the performance, the final version of \algname does not generate actions auto-regressively.

\begin{table}
\begin{minipage}{1.0\linewidth}
\centering
\setlength\tabcolsep{2.0pt}
\scriptsize
\begin{tabular}{@{}lcccccccccccc@{}}
\toprule
 & & & \multicolumn{4}{c}{Distractors} & Backgrounds\\
\cmidrule(lr){4-7} \cmidrule(lr){8-8}
Model & Seen Tasks & Unseen Tasks & All & Easy & Medium & Hard & All &  Inference Time (ms)  \\
\midrule
Gato~\citep{reed2022generalist} & 65 (-32) & 52 (-24) & 43 (-40) & 71 & 44 & 29 & 35 (-24) &  129 \\
BC-Z~\citep{jang2022bc} & 72 (-25) & 19 (-57) & 47 (-36) & 100 & 67 & 7 & 41 (-18) & 5.3 \\
BC-Z XL & 56 (-41) & 43 (-33) & 23 (-60) & 57 & 33 & 0 & 35 (-24) & 5.9 \\
\algname (ours) & \textbf{97} & \textbf{76} & \textbf{83} & 100 & 100 & 64 & \textbf{59} & 15 \\
\midrule
\algname w/o big model & 89 (-8) & 62 (-14) & 77 (-6) & 100 & 100 & 50 & 53 (-6) & 13.5 \\
\algname w/o pre-training & 84 (-13) & 43 (-33) & 60 (-23) & 100 & 67 & 36 & 41 (-18) & 15 \\
\algname w/ continuous actions & 68 (-29) & 43 (-33) & 37 (-46) & 71 & 67 & 0 & 35 (-24) & 16 \\
\algname w/ auto-regressive actions & 85 (-12) & 71 (-5) & 67 (-16) & 100 & 78 & 43 & \textbf{65 (+6)} & 36 \\
\algname w/o history & 82 (-15) & 62 (-14) & 50 (-33) & 71 & 89 & 14 & \textbf{59 (+0)} & 15 \\
\algname w/o Transformer & 86 (-13) & 62 (-14) & 67 (-16) & 100 & 100 & 29 & \textbf{59 (+0)} & 26  \\
\bottomrule
\end{tabular}
	\end{minipage}\hfill
	\begin{minipage}{0.7\linewidth}
	\vspace{0.5cm}
        \centering
        \includegraphics[width=\linewidth]{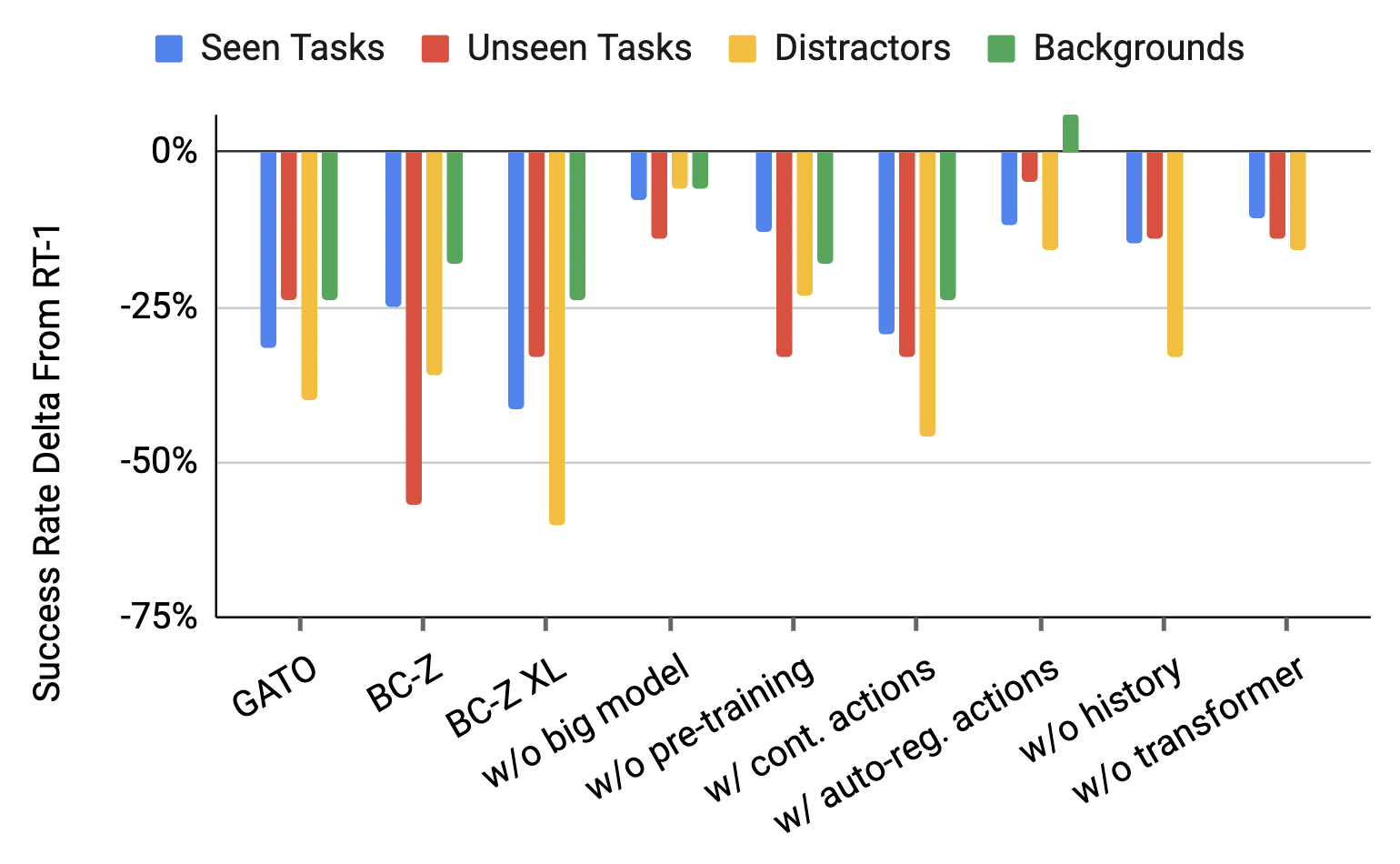}
        \caption{Various model ablations of \algname across seen tasks, generalization to unseen tasks, and robustness to distractors and backgrounds.}
        \label{table:model_ablation}
	\end{minipage}
\end{table}

\subsection{Summary and Analysis}
\label{sec:app_analysis}

In this section, we summarize some of our findings and propose intuition for \algname's high performance, generalization, and robustness. 
First, ImageNet pretraining (along with Universal Sentence Encoder language embedding) has a large impact particularly on unseen tasks.
We observe that \algname inherits some of the knowledge that results from the generality and diversity of the datasets these models were trained on.
Second, continuous actions have a large impact across all aspects of performance. 
This has been previously observed and may be due to
the ability to represent more complex action distributions -- the per-dimension discretization allows our model to represent complex multi-modal distributions, while the Gaussian distribution captures only a single mode.
Third, given such expressive multitask models, data diversity has a larger impact than data size.
Indeed, even datasets collected in simulated environments or from different robotic embodiments can be leveraged by \algname, opening avenues for new regimes of data collection.

\begin{figure}[h!]
    \centering
    \includegraphics[width=\linewidth]{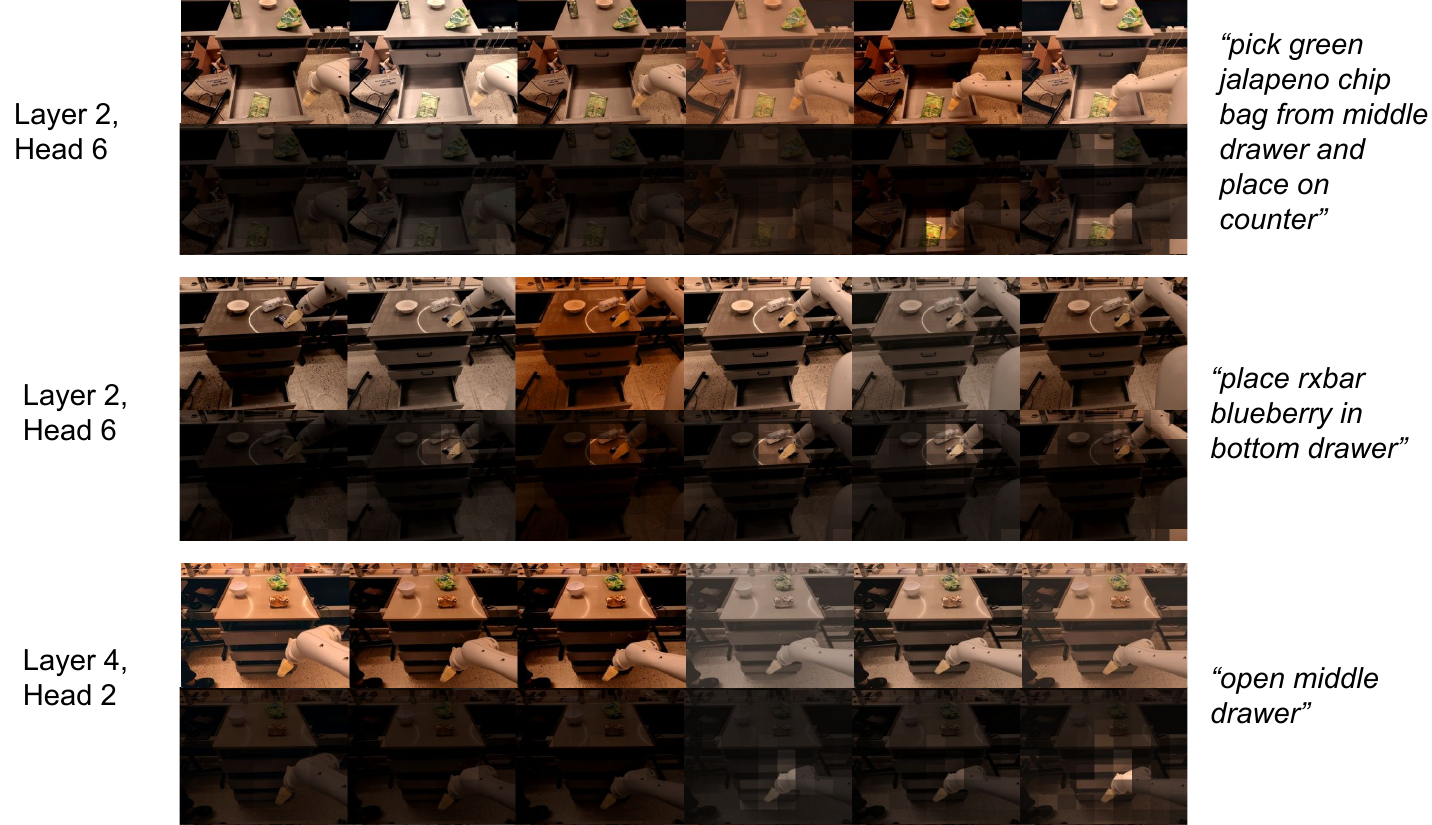}
    \caption{In this figure we show the attention map of the RT-1 policy. Different layers and heads generally focus on different part of the image. Most commonly, they focus on the parts of the scene with the richest interaction affordances, such as graspable objets. For example, Layer 2 Head 6 focuses on the jalapeno chips and pepsi can in grasping tasks; and Layer 4 Head 2 focuses on the drawer in drawer opening tasks.}
    \label{fig:attention_map}
\end{figure}

Finally, \algname fuses language into the image pipeline early via FiLM conditioning, compared to e.g., Gato's late fusion. This enables image tokens that focus only on relevant features for the instruction at hand, which may be the cause of poor distractor performance for Gato.
Figure~\ref{fig:attention_map} visualizes the attention during rollouts of \algname. 
We see that the attention is focused on relevant features and particularly on interaction between the gripper and the object of interest. 
The bottleneck of attention layers such as these results in a compact representation which effectively ignores distractors and varying backgrounds.

\end{document}